\documentclass[lettersize,journal]{IEEEtran}
\usepackage{amsmath,amsfonts}
\usepackage{bbding}
\usepackage{algorithmic}
\usepackage{algorithm}
\usepackage{array}
\usepackage[caption=false,font=footnotesize,labelfont=rm,textfont=rm]{subfig}
\usepackage{textcomp}
\usepackage{stfloats}
\usepackage{url}
\usepackage{verbatim}
\usepackage{graphicx}
\usepackage{cite}
\usepackage{multirow}
\usepackage{multicol}
\usepackage{booktabs}
\usepackage{upgreek}
\usepackage[colorlinks,
            linkcolor=blue,
            anchorcolor=blue,
            urlcolor=blue,
            citecolor=blue]{hyperref}

\hypersetup{colorlinks=true,linkcolor=blue,citecolor=blue}
\begin{document}

\title{MIPD: A Multi-sensory Interactive Perception Dataset for Embodied Intelligent Driving}

\author{Zhiwei Li$^{\scriptscriptstyle }$,
  Tingzhen Zhang, 
  Meihua Zhou, 
  Dandan Tang, 
  Pengwei Zhang, 
  Wenzhuo Liu,
  Qiaoning Yang$^{\scriptscriptstyle }$,
  Tianyu Shen,
  Kunfeng Wang, 
  and Huaping Liu

\thanks{Manuscript received November 8, 2024; revised November 8, 2024. This work was supported in part by the Science and Technology on Metrology and Calibration Laboratory under Grant JLKG2023001B004 and in part by the National Natural Science Foundation of China under Grant 62302047.  
\textit{((Tingzhen Zhang and Zhiwei Li contributed equally to this work.) (Corresponding authors: Qiaoning Yang (yangqn@buct.edu.cn))}}
  \thanks{Zhiwei Li, Tingzhen Zhang, Tianyu Shen, Qiaoning Yang, Kunfeng Wang are with the College of Information Science and Technology, Beijing University of Chemical Technology, Beijing 100029, China.}
  \thanks{Meihua Zhou is with the Institute of Ophthalmology, Beijing Tongren Hospital, Beijing 100730, China.}
  \thanks{Dandan Tang is with the College of Information Science and Engineering, Yanshan University, Hebei 066004, China.}
  \thanks{Pengwei Zhang is with the College of Artificial Intelligence, University of Chinese Academy of Sciences, Beijing 100040, China.}
  \thanks{Wenzhuo Liu is with the College of Mechanics and Vehicles, Beijing Institute of Technology, Beijing 100081, China.}
  \thanks{Huaping Liu is with the College of Computer Science and Technology, Tsinghua University, Beijing 100084, China.}
}

\maketitle

\begin{abstract}
During the process of driving, humans usually rely on multiple senses to gather information and make decisions. Analogously, in order to achieve embodied intelligence in autonomous driving, it is essential to integrate multidimensional sensory information in order to facilitate interaction with the environment. However, the current multi-modal fusion sensing schemes often neglect these additional sensory inputs, hindering the realization of fully autonomous driving. This paper considers multi-sensory information and proposes a multi-modal interactive perception dataset named ParallelBody, enabling expanding the current autonomous driving algorithm framework, for supporting the research on embodied intelligent driving. In addition to the conventional camera, lidar, and 4D radar data, our ParallelBody dataset incorporates multiple sensor inputs including sound, light intensity, vibration intensity and vehicle speed to enrich the dataset comprehensiveness. Comprising 126 consecutive sequences, many exceeding twenty seconds, ParallelBody dataset features over 8,500 meticulously synchronized and annotated frames. Moreover, it encompasses many challenging scenarios, covering various road and lighting conditions. The dataset has undergone thorough experimental validation, producing valuable insights for the exploration of next-generation autonomous driving frameworks. Data, development kit and more details will be available at \url{https://github.com/BUCT-IUSRC/Dataset__MIPD}.

\end{abstract}


\begin{figure}[!ht]
\centering
\subfloat[Acquisition Platform]{\includegraphics[width=1\linewidth]{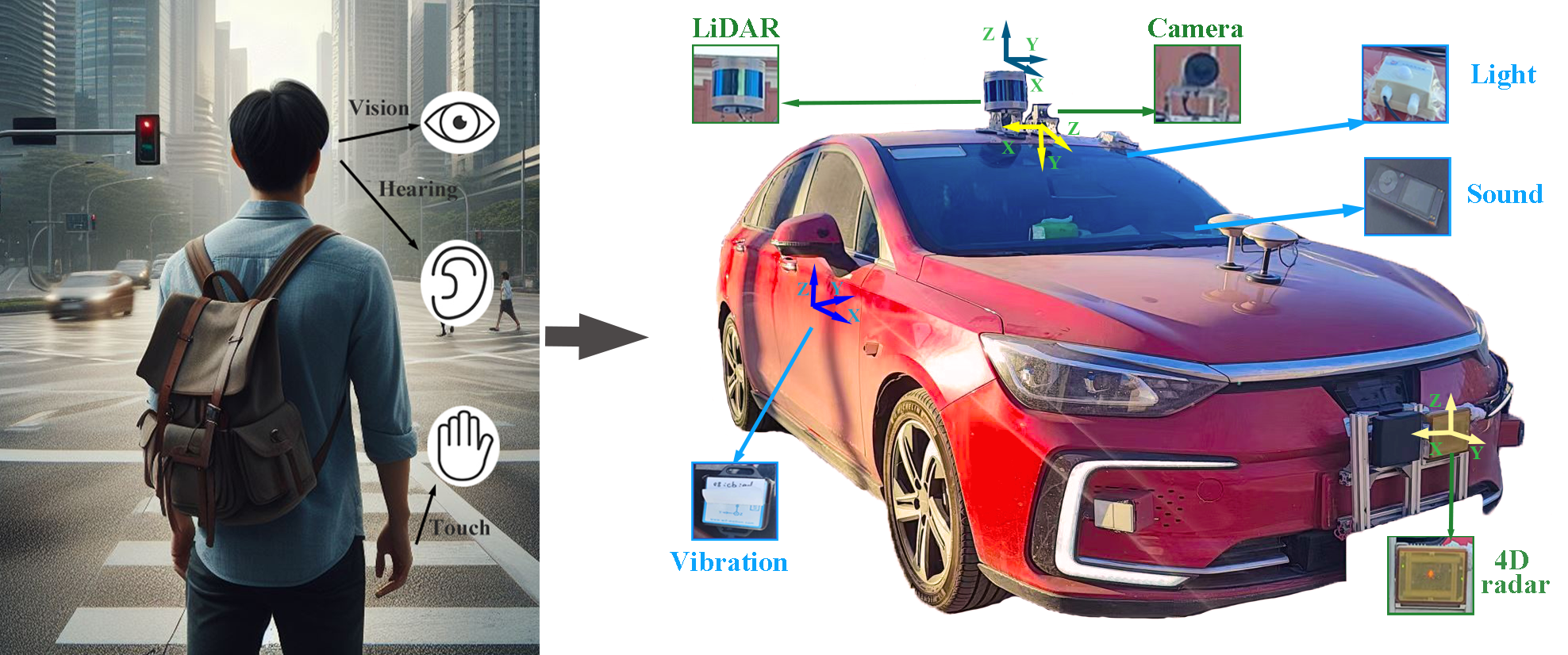}
\label{people+car}}\\

\subfloat[Camera]{\includegraphics[width=0.5\linewidth, height=2.8cm]{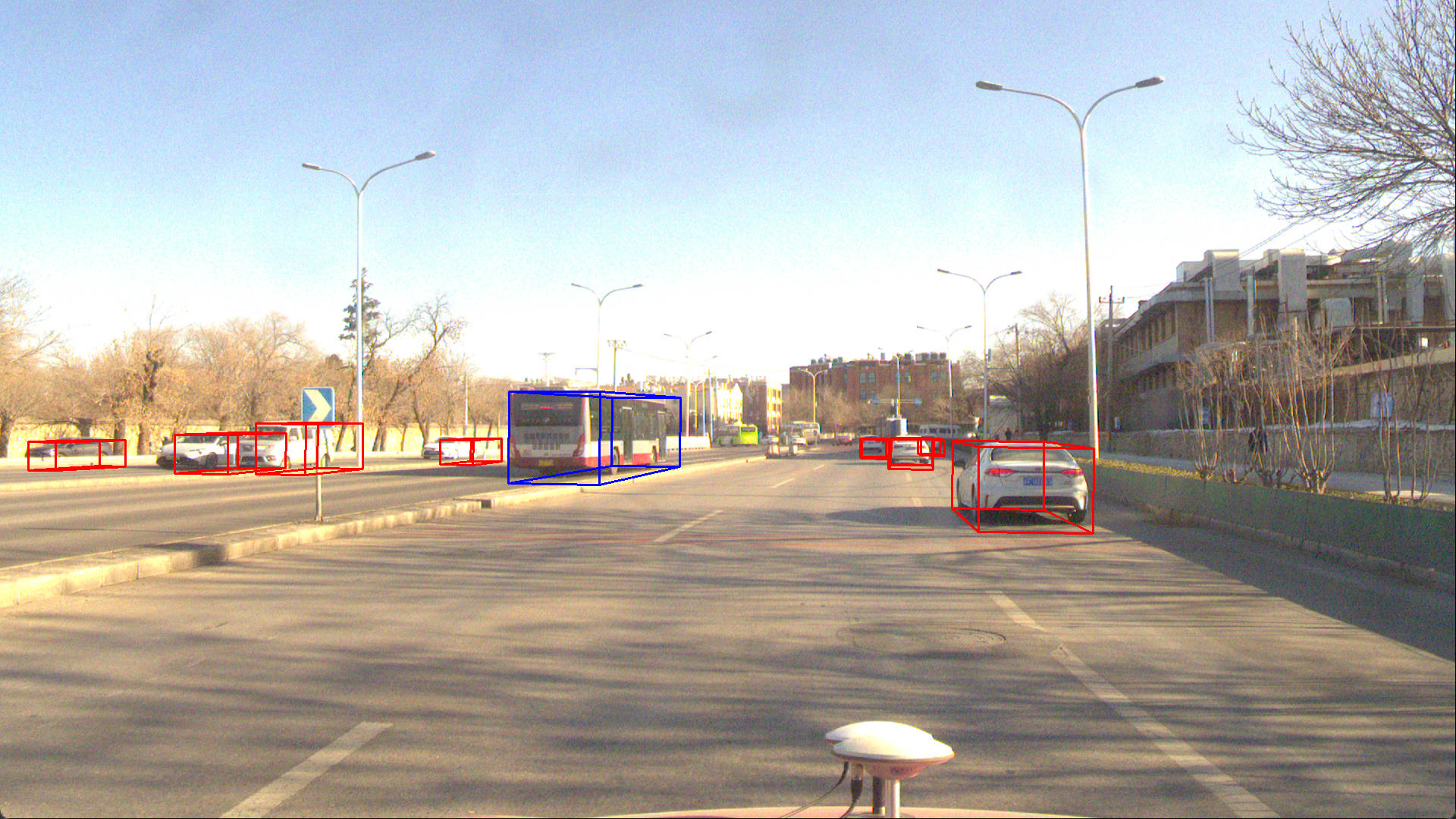}
\label{img}}
\subfloat[LiDAR]{\includegraphics[width=0.5\linewidth, height=2.8cm]{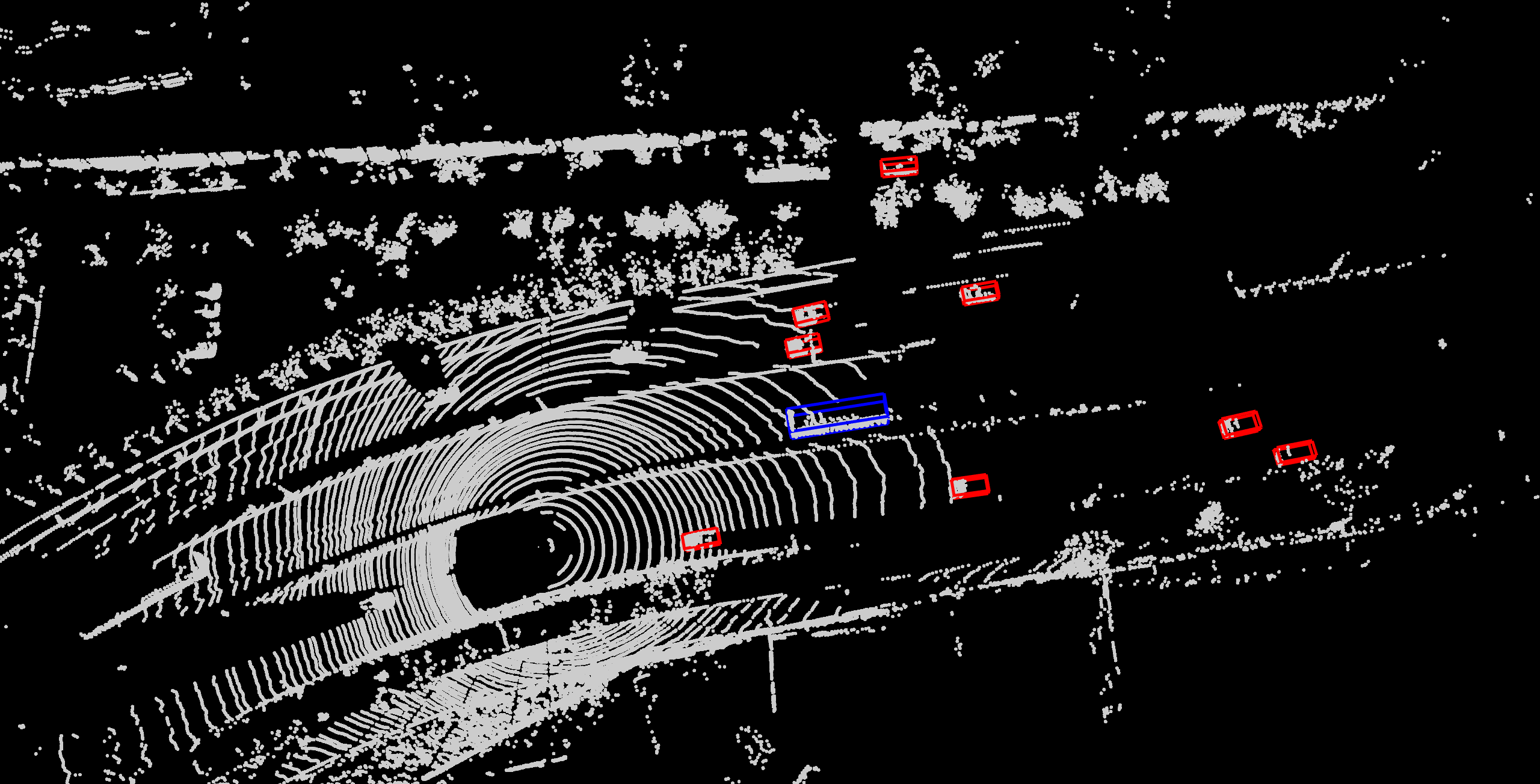}
\label{lidar}}\\

\subfloat[4D radar]{\includegraphics[width=0.5\linewidth, height=2.8cm]{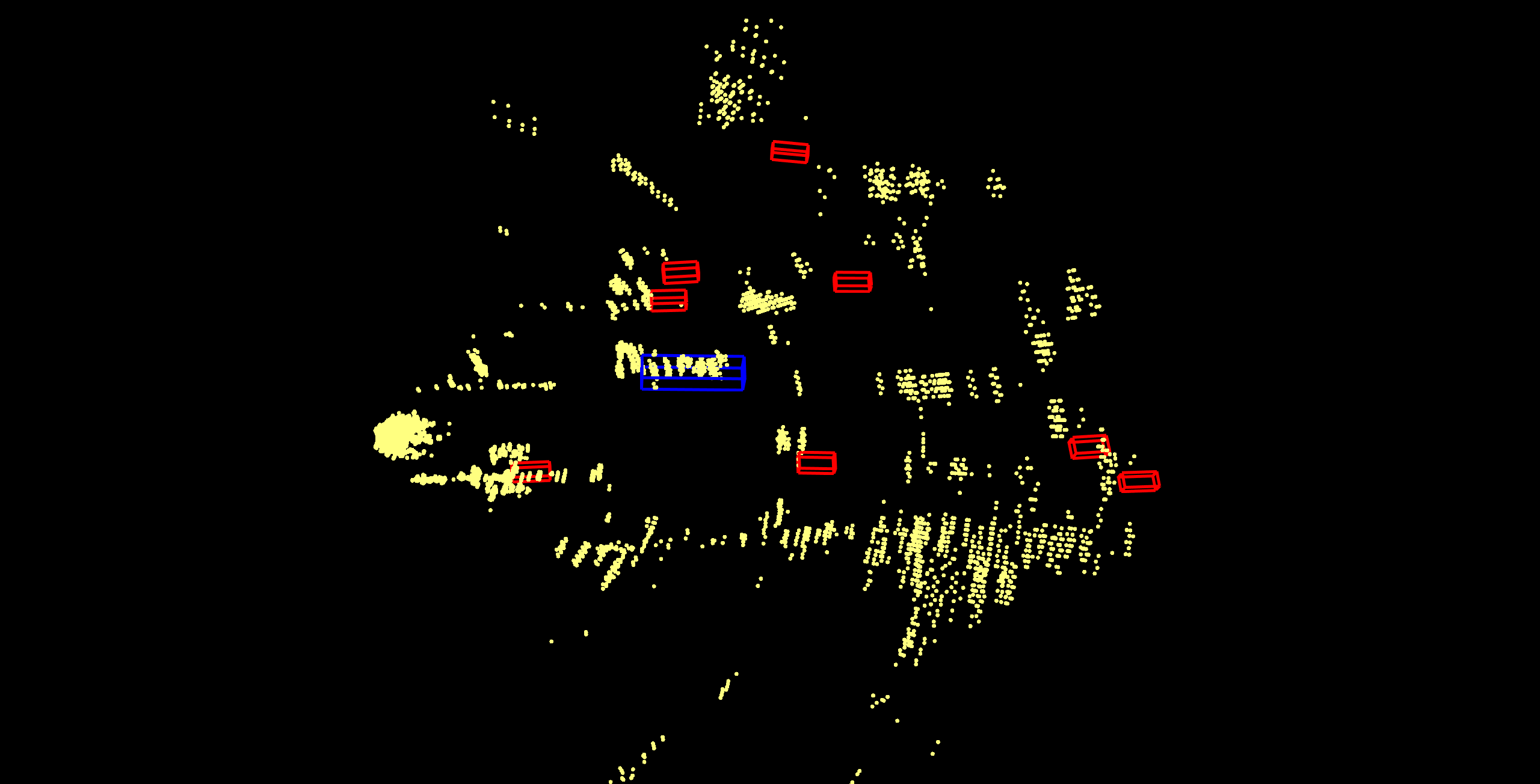}
\label{arbe}}
\subfloat[Sound]{\includegraphics[width=0.5\linewidth, height=2.8cm]{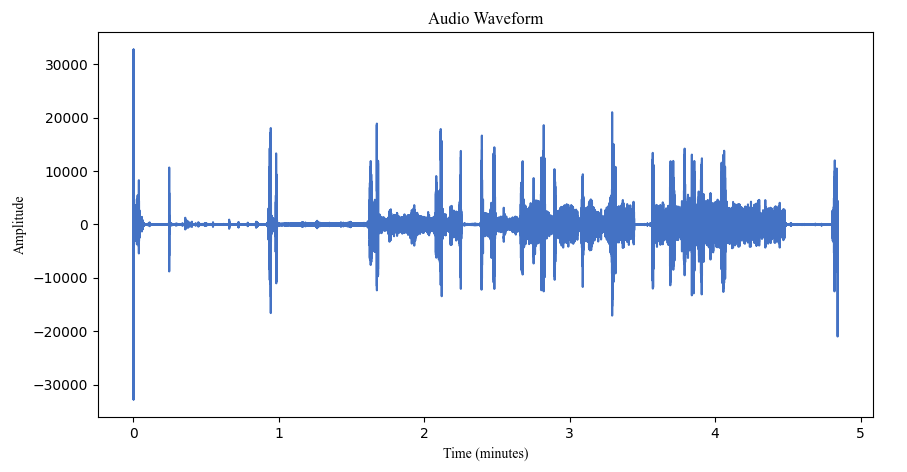}
\label{sound}}\\

\subfloat[Light]{\includegraphics[width=0.5\linewidth, height=2.8cm]{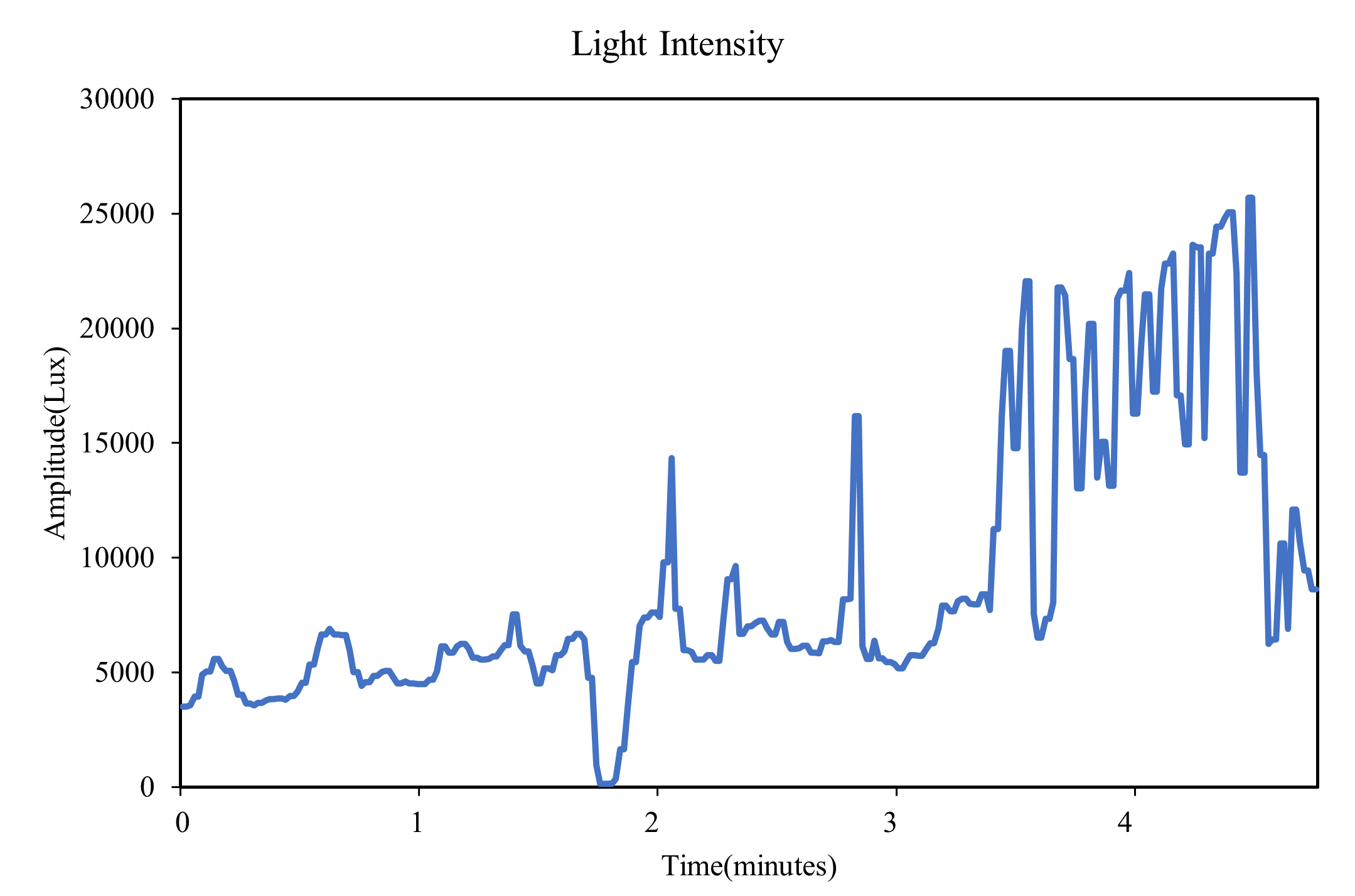}
\label{light}}
\subfloat[Vibration]{\includegraphics[width=0.5\linewidth, height=2.8cm]{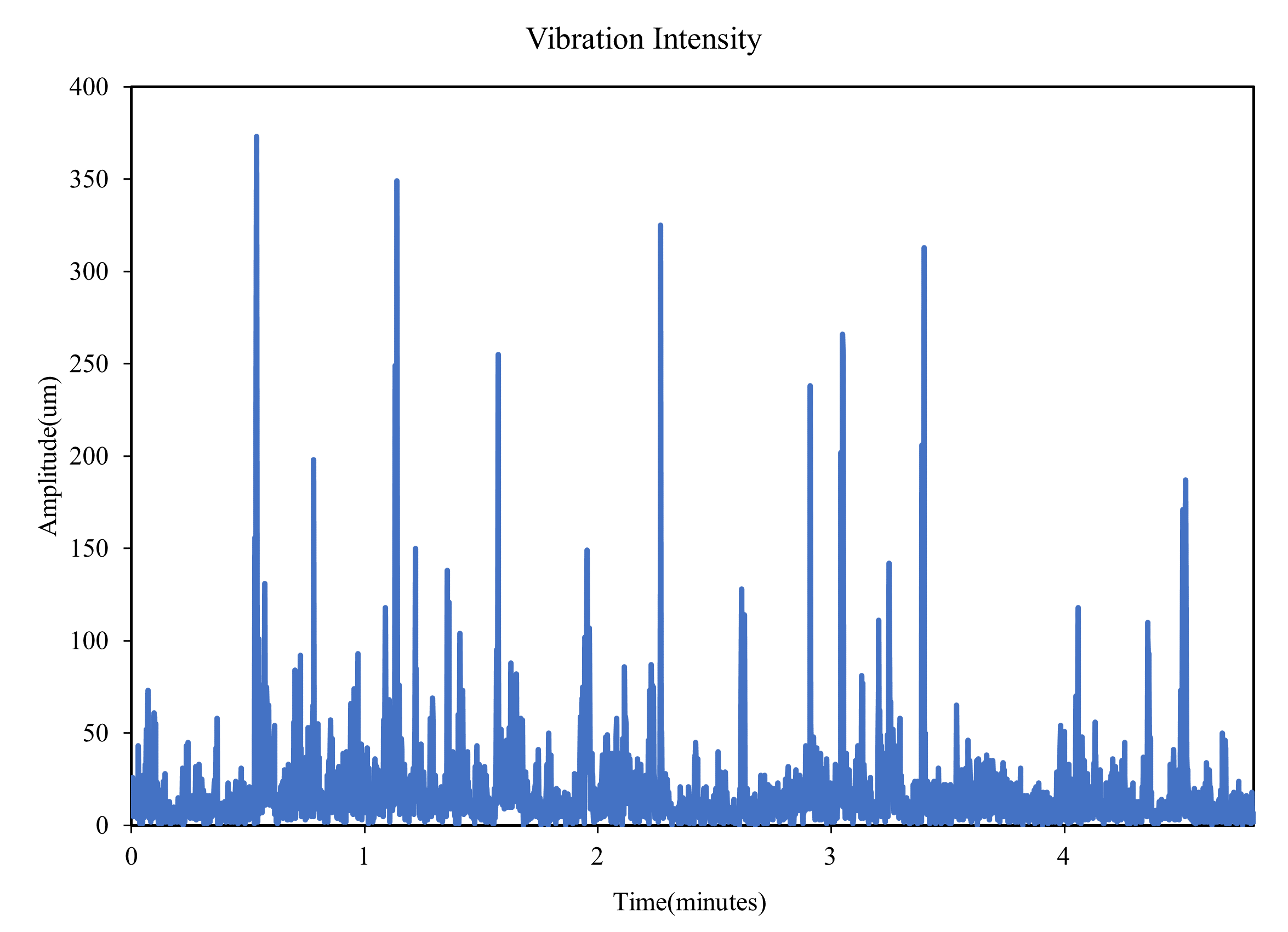}
\label{vibration}}\\

\caption{The configuration of our experiment platform and visualization scenarios on the data collected by different sensors. (a) shows information of each sensor coordinate system in the data acquisition platform. (b), (c), (d), (e), (f), (g) shows the results after visualizing our data.}
\label{fig_1}
\end{figure}

\section{Introduction}
\IEEEPARstart{T}{h}e evolution of autonomous driving technology heralds a transformative era in transportation systems, where the depth and integration of environmental perception are central to enhancing road safety, easing traffic pressures, and optimizing energy utilization\cite{shen0}. While autonomous driving has made commendable strides towards fully autonomous operations, its perceptual limitations continue to pose significant challenges for its broader adoption\cite{jl2011}. Thus, the incorporation of embodied intelligence into autonomous driving is critical, with interaction perception being a fundamental and pivotal concept. Interaction perception underscores the potential of multi-modal sensory data to enhance the safe and efficient interaction of vehicles with other traffic participants, enabling the acquisition and interpretation of environmental information.

In the concept of interaction perception, vehicles not only passively receive information from their surroundings but also influence the environment and other traffic participants through their own behavior. This two-way interaction involves the acquisition, processing, and delivery of multidimensional sensory data. For example, interactive perception may include the vehicle receiving sensor data from its surroundings, including images, point clouds, light, and vibration. Furthermore, the vehicle is capable of influencing its surroundings and other traffic participants based on information about its own state, including data such as acceleration, steering angle, and so forth. It is possible for vehicles to predict the impact and to dynamically optimize and adjust their perceptual state based on their own state.

The construction of a multidimensional perceptual dataset is a fundamental step in the realization of the goal of interactive perception. Notably, the deficiencies in current datasets to accurately capture and adapt to the dynamic variations of the real world represent formidable barriers to further advancements\cite{zs2024,kw2023}. These datasets often depend excessively on conventional visual and distance-sensing technologies like cameras, LiDAR, and radar\cite{li1,bt2020,jl2023,shi2023bssnet,wang2023path}. Prominent datasets such as Kitti\cite{kitti}, NuScenes\cite{nuScenes}, Waymo\cite{waymo}, and Argoverse\cite{argoverse}, although abundant in visual data, fall short in offering a holistic perception of the environment. This oversight limits autonomous systems from fully understanding the nuances of environmental changes, particularly the subtle variables like variations in lighting conditions across different weather scenarios or the differing responses of vehicles to varied road surfaces\cite{lw2023,xz2023}. Such limitations starkly curtail the systems' capability to comprehend and adapt to intricate environments\cite{li2}.

Furthermore, the complexity and variability of real-world driving conditions far exceed the scope of existing datasets, encompassing a spectrum of dynamic elements such as light intensity and road conditions\cite{ll2020,ib2017}. The existing data set is inadequate to meet the needs of autonomous driving technologies, which require a more comprehensive and dynamic compilation of environmental data\cite{jg2019}. This is essential for enhancing the adaptability and decision-making precision of autonomous driving technologies.

In light of these challenges, an innovative multi-modal dataset design concept is introduced in this paper to address existing deficiencies. Considering that in complex transportation scenarios, people tend to perceive dynamic elements comprehensively through visual, auditory, and tactile multi-sensory synergies, the data collection platform was extended as shown in Fig. \ref{fig_1}\subref{people+car}. Our proposed dataset goes beyond conventional visual data to include camera imagery, lidar and 4D radar data, and importantly, it integrates diverse multi-dimensional information such as vibration and light intensity. This robust amalgamation of data aims to offer a more accurate and holistic environmental simulation, significantly bolstering the adaptive capabilities and situational awareness of autonomous driving systems.

Our main contributions to this work are set out below:

(1) A novel multi-modal dataset is proposed and constructed, integrating a variety of information such as camera data, point cloud information from lidar, 4D radar, sound, vibration, light intensity, and vehicle speed. This dataset is designed to comprehensively enhance perception tasks by fusing multi-sensory data to compensate for the lack of single-sensory information.

(2) Our dataset consists of 126 consecutive sequences, most of which last more than twenty seconds and contain over 8,500 carefully synchronized and annotated frames. Additionally, our dataset includes challenging scenarios, such as various road conditions, weather conditions, lighting conditions, and more.

(3) Experiments were conducted using multiple single-modal and multi-modal correlation models to validate the effectiveness of the collected dataset.

The paper is organized as follows: In Section 2, a summary of previous research on different sensor configurations of multi-modal datasets, fusion perception models, and embodied perception is provided. Section 3 presents detailed information about the dataset, including sensor setup details, dataset labeling and visualization, and statistical analysis. The experiment details and the discussion of the results are presented in Section 4. In Section 5, the reasons for the significant degradation of the perception accuracy of existing perceptual models when confronted with complex scenarios are discussed. Furthermore, the limitations of the existing algorithms are analyzed and potential future research directions are outlined. Finally, in Section 6, a summary of the work is provided and the limitations of existing datasets in comparison to the strengths of the dataset are discussed.

\section{related work}
\subsection{Multi-modal Datasets}
KITTI\cite{kitti}, NuScenes\cite{nuScenes}, Waymo\cite{waymo}, and Argoverse\cite{argoverse} are widely used autonomous driving multi-modal datasets that are favored for their large driving scenarios and sample sizes, as well as high quality annotation information. The KITTI dataset is the first to employ a Front-view Camera, LiDAR, and Inertial Measurement Unit (IMU) as sensors to support multiple tasks such as target detection, target tracking, depth estimation, and more. The camera has high robustness, but is affected by light and weather conditions, in low light, backlight or rain, the camera image lacks RGB information to observe the object. In these scenes, LiDAR provides dense spatial information that makes up for the lack of cameras. However, in scenes with more objects, LiDAR cannot effectively distinguish between overlapping or close objects. Inertial measurement units are sensors that detect and measure acceleration, tilt, shock, vibration, rotation, and multi-degree of freedom (DoF) motion and are an important part of navigation, orientation, and motion control. Datasets such as NuScenes, Waymo, and Argoverse are equipped with round-view cameras that support multiple modular tasks such as perception, decision making, and planning due to the rich scale of the scene. The NuScenes dataset also introduces mapping data for decision-making and planning tasks for the first time. In addition, the data of CAN-bus, GPS, HR Camera, Radar sensor and other sensors provide rich data information for the research of perception algorithms and multi-modal fusion algorithms under different types of sensors. The configurations for different datasets are shown in TABLE \ref{tab:datasets} and Table \ref{tab:datasats2}.
\begin{table*}[h!t]
\caption{SENSOR CONFIGURATION FOR CURRENT MAINSTREAM AUTONOMOUS DRIVING DATASETS\label{tab:datasets} }
\centering
\resizebox{\linewidth}{!}{
\footnotesize
\renewcommand{\arraystretch}{1.2}
\begin{tabular}{c c c c c c c c c c c c c c}
\toprule[2px]
\multirow{2}*{\textbf{Dataset}} & \multirow{2}*{\textbf{Year}} & \multicolumn{12}{c}{\textbf{Sensor type}} \\ \cline{3-14}
 &  & \textbf{Front-view Camera} & \textbf{360°} & \textbf{LiDAR} & \textbf{Radar}  & \textbf{CAN-bus} & \textbf{GPS} & \textbf{IMU} & \textbf{Sound} & \textbf{HDMap} & \textbf{HRCamera}& \textbf{Light} & \textbf{Vibration} \\ \hline
KITTI\cite{kitti} & 2012 & \Checkmark &\XSolidBrush & \Checkmark & \XSolidBrush & \XSolidBrush& \Checkmark& \Checkmark &\XSolidBrush &\XSolidBrush & \XSolidBrush& \XSolidBrush& \XSolidBrush\\ 

Apolloscape\cite{apolloscape} & 2016 & \Checkmark & \XSolidBrush & \XSolidBrush &\XSolidBrush &\XSolidBrush & \Checkmark&\Checkmark &\XSolidBrush &\XSolidBrush &\XSolidBrush & \XSolidBrush& \XSolidBrush\\ 

Woodscape\cite{woodscape} & 2019 & \XSolidBrush & \Checkmark & \Checkmark & \XSolidBrush  &\Checkmark & \Checkmark&\Checkmark & \XSolidBrush& \XSolidBrush&\XSolidBrush & \XSolidBrush& \XSolidBrush\\ 

Nuscenes\cite{nuScenes} & 2019 & \XSolidBrush& \Checkmark& \Checkmark & \Checkmark &   \Checkmark & \Checkmark &\XSolidBrush &\XSolidBrush & \Checkmark & \XSolidBrush& \XSolidBrush& \XSolidBrush\\

Waymo\cite{waymo}& 2019 &\XSolidBrush & \Checkmark & \Checkmark & \XSolidBrush&\XSolidBrush &\XSolidBrush  &\XSolidBrush &\XSolidBrush & \XSolidBrush&\XSolidBrush & \XSolidBrush& \XSolidBrush\\

Argoverse\cite{argoverse}& 2019 & \XSolidBrush& \Checkmark & \Checkmark & \XSolidBrush& \XSolidBrush & \XSolidBrush&\XSolidBrush &\XSolidBrush & \Checkmark & \XSolidBrush& \XSolidBrush& \XSolidBrush\\

CARRAD\cite{carrad} & 2020 & \Checkmark& \XSolidBrush&\XSolidBrush  & \Checkmark &\XSolidBrush & \XSolidBrush & \XSolidBrush&\XSolidBrush & \XSolidBrush&\XSolidBrush & \XSolidBrush & \XSolidBrush \\

ONCE\cite{once}& 2021 & \XSolidBrush& \Checkmark & \Checkmark &\XSolidBrush &\XSolidBrush &\XSolidBrush &\XSolidBrush & \XSolidBrush&\XSolidBrush &\XSolidBrush & \XSolidBrush& \XSolidBrush\\

ZOD\cite{zod}  & 2023 &\XSolidBrush & \XSolidBrush& \Checkmark & \XSolidBrush& \XSolidBrush & \XSolidBrush& \Checkmark &\XSolidBrush &\XSolidBrush & \Checkmark & \XSolidBrush& \XSolidBrush\\

Ours  & 2024 &\Checkmark & \XSolidBrush& \Checkmark & \Checkmark& \XSolidBrush & \XSolidBrush& \Checkmark & \Checkmark &\XSolidBrush &\XSolidBrush  & \Checkmark& \Checkmark\\

\bottomrule[2px]
\end{tabular}
}
\end{table*}
\begin{table*}[h!t]
  \begin{center}
    \caption{THE OVERVIEW OF THE AUTONOMOUS DRIVING DATASETS FOR OBJECT DETECTION AND TRACKING\label{tab:datasats2}}
    \centering
    \setlength{\tabcolsep}{11pt}
    \renewcommand{\arraystretch}{1.2}
    \begin{tabular}{c c c c c c c c c c}
    
    \toprule[2px]

      \multirow{2}{*}{\textbf{Dataset}} & \multirow{2}{*}{\textbf{Year}} & \multicolumn{5}{c}{\textbf{Scenarios}} & \multirow{2}{*}{\textbf{Light}} &\multicolumn{2}{c}{\textbf{Annotations}}\\ \cline{3-7} \cline{9-10}
      \multirow{2}{*}{} & \multirow{2}{*}{}& \textbf{Urban} & \textbf{Campus} & \textbf{Highway} &\textbf{Tunnel}& \textbf{Parking} &\multirow{2}{*}{} & \textbf{3D BOX}& \textbf{Track ID} \\ \hline
      KITTI\cite{kitti}       & 2012 & \Checkmark & \Checkmark & \Checkmark & \XSolidBrush & \Checkmark & \Checkmark & \Checkmark & \XSolidBrush \\
Apolloscape\cite{apolloscape} & 2016 & \Checkmark & \XSolidBrush & \Checkmark & \XSolidBrush & \Checkmark & \Checkmark & \Checkmark & \Checkmark \\
Woodscape\cite{woodscape}   & 2019 & \Checkmark & \XSolidBrush & \Checkmark & \XSolidBrush & \Checkmark & \XSolidBrush & \Checkmark & \XSolidBrush \\
Nuscenes\cite{nuScenes}    & 2019 & \Checkmark & \XSolidBrush & \XSolidBrush & \XSolidBrush & \Checkmark & \Checkmark & \Checkmark & \Checkmark \\
Waymo\cite{waymo}       & 2019 & \Checkmark & \XSolidBrush & \Checkmark & \XSolidBrush & \XSolidBrush & \Checkmark & \Checkmark & \Checkmark \\
Argoverse\cite{argoverse}   & 2019 & \Checkmark & \XSolidBrush & \XSolidBrush & \XSolidBrush & \XSolidBrush & \Checkmark & \Checkmark & \Checkmark \\
CARRAD\cite{carrad}      & 2020 & \Checkmark & \XSolidBrush & \XSolidBrush & \XSolidBrush & \Checkmark & \XSolidBrush & \Checkmark & \Checkmark \\
ONCE\cite{once}        & 2021 & \Checkmark & \XSolidBrush & \Checkmark & \Checkmark & \XSolidBrush & \Checkmark & \Checkmark & \XSolidBrush \\
ZOD\cite{zod}         & 2023 & \Checkmark & \XSolidBrush & \Checkmark & \XSolidBrush & \XSolidBrush & \Checkmark & \Checkmark & \XSolidBrush \\
Ours         & 2024 & \Checkmark & \Checkmark & \Checkmark & \XSolidBrush & \XSolidBrush & \Checkmark & \Checkmark & \Checkmark \\
    
    \bottomrule[2px]    
    \end{tabular}
  \end{center}
\end{table*}

\subsection{Fusion Perception}
The fusion of multiple sensors can compensate for the limitations of a single sensor, thereby providing more comprehensive and accurate perceptual results. \cite{xz2023,gong2022multi}
In recent years, numerous studies have been conducted by researchers on perceptual algorithms that integrate multi-modal data. Among these, lidar and camera image features naturally contain complementary information, and the fusion perception algorithm for these two modes is currently the most mainstream algorithm. For instance, Li et al. \cite{YL2022} employ InverseAug to reverse geometric correlation enhancement, thereby achieving accurate geometric alignment of lidar points and image pixels. Additionally, they utilize a dynamic correlation between images and lidar features during the fusion process, which ultimately leads to more accurate perception results. The DeepInteraction architecture proposed by \cite{ZY2022} represents a significant advance over existing multi-modal fusion strategies. It employs a multi-modal representational interactive encoder and a multi-modal predictive interactive decoder to learn and maintain unique representations of each mode. End-to-end multi-modal 3D target detection was achieved by Yan et al. \cite{JY2023} through the use of positional encoding of multi-view images and point clouds, combined with the addition of corresponding modal markers, thereby eliminating the repeated projection and sampling processes.

Although the advantages of cameras and LiDAR can be complementary, the quality of information captured by both of them is significantly reduced when faced with adverse weather conditions, such as rain, which significantly reduces the detection accuracy \cite{YY2019}. Consequently, in recent years, researchers have also begun to explore the potential of sensing algorithms that fuse millimeter wave radar and cameras. For instance, Zhang and others \cite{zhang2023dual} published the inaugural multi-modal autonomous driving dataset comprising diverse 4D radar data, and empirically demonstrated the necessity of radar data in perception tasks.

In addition to the aforementioned sensors, recent scholarship has also highlighted the significance of one-dimensional information, such as vehicle speed, steering angle, and IMU data, for the automatic driving perception task \cite{liu2024fmdnet,gong2023sifdrivenet}. For instance, Gong et al. \cite{Y.Gong2024} proposed a lane line detection method that fuses monocular image and vibration information. The method utilizes the vibration signals generated when the vehicle passes the vibration markings as supervisory information for lane occlusion prediction. It provides occlusion a priori knowledge for the image lane line detection network and adaptively adjusts the network weights to improve the detection performance. However, the algorithm in question only makes use of image and vibration data, and thus lacks the utilisation of other modal data. The lack of one-dimensional environmental state data, such as vibration and light, in the automated driving multi-modal dataset results in the existing automated driving perception algorithms being unable to effectively utilise environmental state information. This hinders the ability to perceive the surrounding environment in a comprehensive and accurate manner, and leads to suboptimal performance in complex open scenarios, such as backlighting and bumpy roads.

\subsection{Embodied Perception}
Embodied intelligence represents a significant advancement in the field of autonomous driving, transcending conventional AI paradigms and emphasizing the intricate interplay between computational systems and the physical world.  A prerequisite for realizing this interaction is the realization of embodied perception.

Embodied perception has been proposed as an important concept and has been widely studied and practiced in the field of robotics. Santhosh K. Ramakrishnan et al. \cite{Ramakrishnan2020} investigated embodied visual exploration in unstructured environments, proposed a classification of existing visual exploration algorithms, and conducted experiments in realistic 3D simulation environments research, providing new performance metrics and benchmarks. Franklin Kenghagho et al. \cite{Kenghagho2022} proposed a new white-box and causal-generative model (NaivPhys4RP) that mimics human perception and explains perceptual problems in complex environments by capturing the five dimensions of functionality, physicality, causality, intention, and utility. In addition, Qianfan Zhao et al. \cite{qzhao2022} proposed a new embodied dataset for robotic active vision learning, which enables researchers to simulate robotic movements and interactions in indoor environments using real point cloud data collected densely in seven real indoor scenarios, thus improving visual performance in novel environments.

In the field of autonomous driving, the realization of embodied perception is based on the construction of multidimensional sensory datasets, which involves the integration of input data from diverse sensors. The construction of detailed environmental models enables autonomous vehicles to make informed decisions and navigate complex traffic scenarios in a safe and efficient manner. Alqudah et al. \cite{alqudah2019audition} demonstrate that the integration of auditory signals significantly boosts a vehicle's responsiveness to dynamic environmental changes, reinforcing the need for multidimensional perceptual information in autonomous driving technologies. Similarly, Wang et al. \cite{wang2021haptic} established that tactile feedback improves the accuracy of driver intention recognition systems, thereby enhancing control and safety in semi-autonomous vehicles, highlighting the critical role of sensory feedback in vehicle automation. 

Embodied intelligence in autonomous driving represents an advanced approach that emphasizes the interaction between computational systems and the physical world. This approach relies on embodied perception, which builds a multidimensional perceptual dataset by fusing data from multiple sensors. Such a dataset not only improves the vehicle's comprehensive understanding of its surroundings, but also significantly enhances the vehicle's responsiveness to dynamic environmental changes and accurate recognition of the driver's intentions, thus improving the overall performance of the autonomous driving system while ensuring safety.

\section{ParallelBody Dataset}
\subsection{Overview of the Dataset}
In contemporary autonomous driving research, the comprehensiveness of multi-modal sensory datasets is critical to advancing perception algorithms. This paper introduces a novel dataset that integrates multi-dimensional information, including vibration data and light intensity, to enhance the perception capabilities of autonomous driving systems in real-world scenarios.

Our dataset provides data collected by several different types of sensors after careful synchronization and annotation, and in the sensor setup, the coordinate relationship between our vehicle and multiple sensors is shown in Figure 1. Our data acquisition system consists of a high-resolution Camera, an 80-line LiDAR, a LiDAR, an IMU, a Light Sensor, four Vibration Sensors, and a Sound Acquisition Sensor. The Camera, LiDAR and Light Sensors are mounted on the roof of the autobahn, the LiDAR is mounted on the front of the vehicle, and the acquisition antennas of the IMU are mounted on the front and rear hood of the autobahn, respectively. Four Vibration Sensors are mounted at each of the four-door locations to fully capture the vibration of the vehicle. Sound Sensors are installed in front of the driver to simulate the sound heard by the driver during real driving. Due to the limited range of horizontal viewing angles of the Camera and the LiDAR, only the data from the front of the vehicle are labeled, and the specific sensor specification parameters are shown in Table \ref{tab:sensors}.

Calibration among various sensors is a crucial task in the field. Numerous researchers have proposed distinct calibration methods tailored to different sensors\cite{jb2022,lc2023,hx2019,gy2023}. Notably, J et al.\cite{jo2018} and H et al.\cite{zh2023} have provided comprehensive summaries and comparisons of various sensor calibration techniques. In our dataset, calibration is performed offline, utilizing a joint LiDAR-Camera calibration and a joint LiDAR-4D Radar calibration, respectively. Specifically, a spherical coordinate system is used to achieve joint calibration between sensors. Tools such as angular reflectors and calibration plates are employed, utilizing the principle of a rigid transformer with spherical 3D information. The system adaptation of the sensors varies, considering the high demand for disk writes for simultaneous data acquisition by multiple sensors. Our camera, LiDAR, radar and IMU data are collected by an industrial computer A with a Linux system, and light sensor, vibration sensor and sound sensor are collected by an industrial computer B with a Windows 11 system. For time synchronization, a method based on the PTP protocol is adopted. Time information is obtained through the IMU and granted to industrial controller A, which in turn grants the time to the camera and LiDAR. Time alignment is performed manually between ICM A and ICM B. The light sensor, vibration sensor and sound sensor are timed by ICM B to ensure time synchronization.

In our dataset, 3D bounding boxes, object labels, and tracking IDs for individual objects are provided for the Camera and LiDAR data. More than a dozen categories are labeled, with a focus on "cars", "pedestrians", "cyclists", "buses", and "trucks", while the remaining objects are categorized as "other". Additionally, light intensity, vibration conditions, and ambient noise are recorded, offering rich auxiliary information for autonomous driving perception algorithms. To ensure data integrity and usability, cues at the beginning, end, and end of each minute were labeled using collector dictation. Based on the statistics of the collected raw data, about 20,000 synchronization frames were extracted and 8,568 of them were annotated. From these annotated frames, 90,983 objects were labeled.

\begin{table*}[h!t]

  \begin{center}
    \caption{THE CONFIGURATION OF OUR AUTONOMOUS VEHICLE SYSTEM PLATFORM\label{tab:sensors}}
    \centering
    \footnotesize
    \setlength{\tabcolsep}{12pt}
    \renewcommand{\arraystretch}{1.5}
    \begin{tabular}{c c c c c c c c c c}
    \toprule[2px]

      \multirow{2}{*}{\textbf{Sensors}} & \multirow{2}{*}{\textbf{Type}} & \multicolumn{3}{c}{\textbf{Resolution}} & &\multicolumn{3}{c}{\textbf{FOV}} & \multirow{2}{*}{\textbf{FPS}}\\ \cline{3-5} \cline{7-9}
      \multirow{2}{*}{} & \multirow{2}{*}{}& \textbf{Range} & \textbf{Azimuth} & \textbf{Elevation} & & \textbf{Range} & \textbf{Azimuth} & \textbf{Elevation}&\multirow{2}{*}{} \\ \hline
      Camera &acA1920-40uc & - & 1920px & 1080px & & - & - & - & 10 \\
      LiDAR & RS-Ruby Lite & 0.05m & $0.2^\circ$ & $0.2^\circ$ & &230m &$360^\circ$ & $40^\circ$ & 10 \\
      4D radar & Arbe Phoenix & 0.3m & $1.25^\circ$ & $2^\circ$ & &153.6m &$100^\circ$ & $30^\circ$ & 20 \\
    \bottomrule[2px]
    \end{tabular}
  \end{center}
\end{table*}

\subsection{Scene Characterization and Classification}
This dataset contains data from two different environments, campus and urban, both of which are extremely challenging in the field of autonomous driving. The campus environment is relatively closed and controlled, providing a simpler test scenario in which traffic is usually more organized. However, vehicle vibration data is relatively abundant due to the high mobility of pedestrians and bicycles on campus, as well as the high number of speed bumps. At the same time, the campus environment is relatively quiet, with no honking allowed in most areas, resulting in relatively little noise. Conversely, the urban environment introduces a more complex and dynamic traffic interaction scenario, characterized by numerous vehicles (buses, cars, trucks), high noise levels, and unpredictable traffic patterns, thus posing greater challenges for autonomous perception systems.

\subsection{Statistical Analysis}
\begin{figure*}[!t]
\centering
\subfloat[]{\includegraphics[width=2.5in]{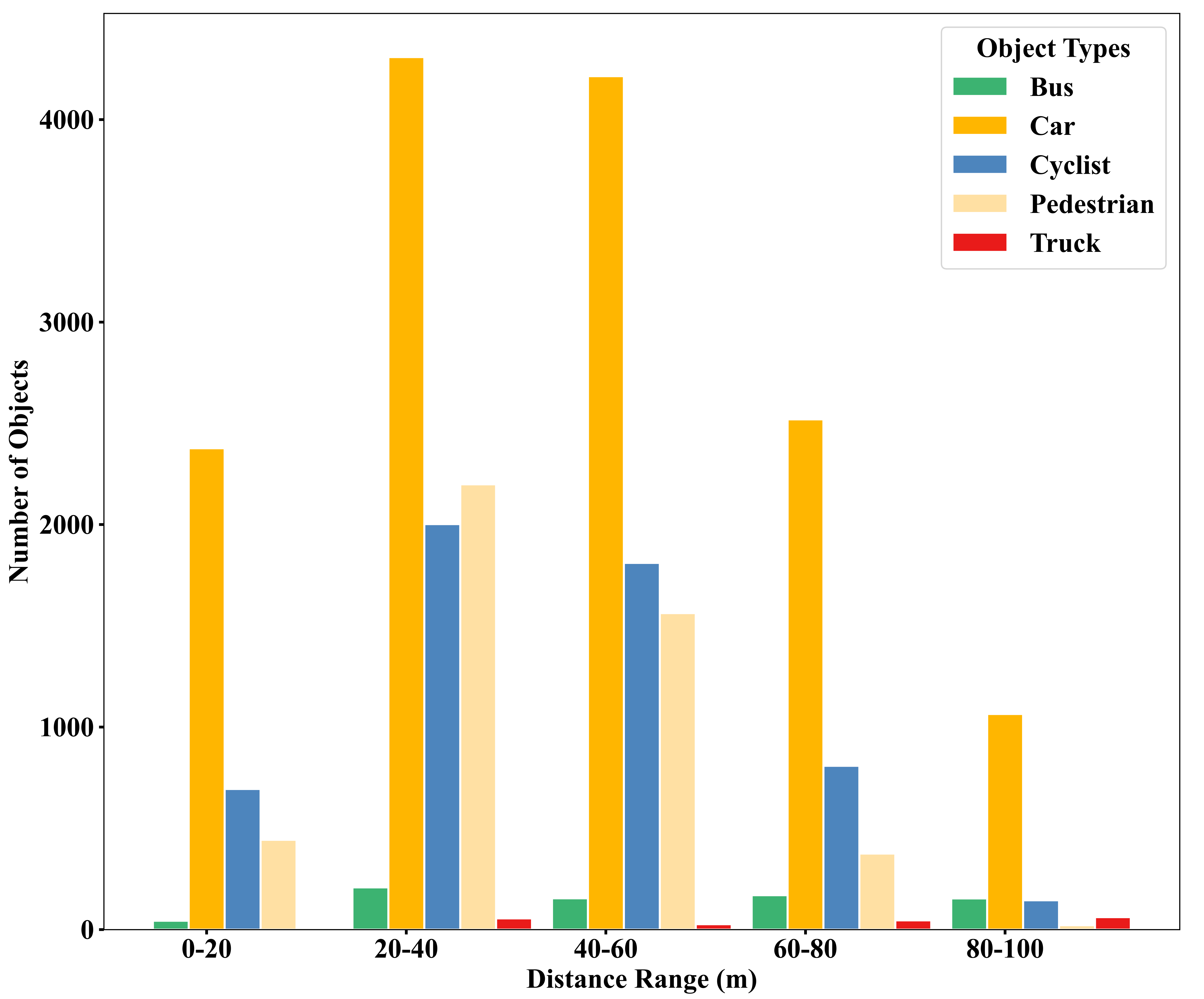}
\label{fig_dis_city}}
\hfil
\subfloat[]{\includegraphics[width=2.5in]{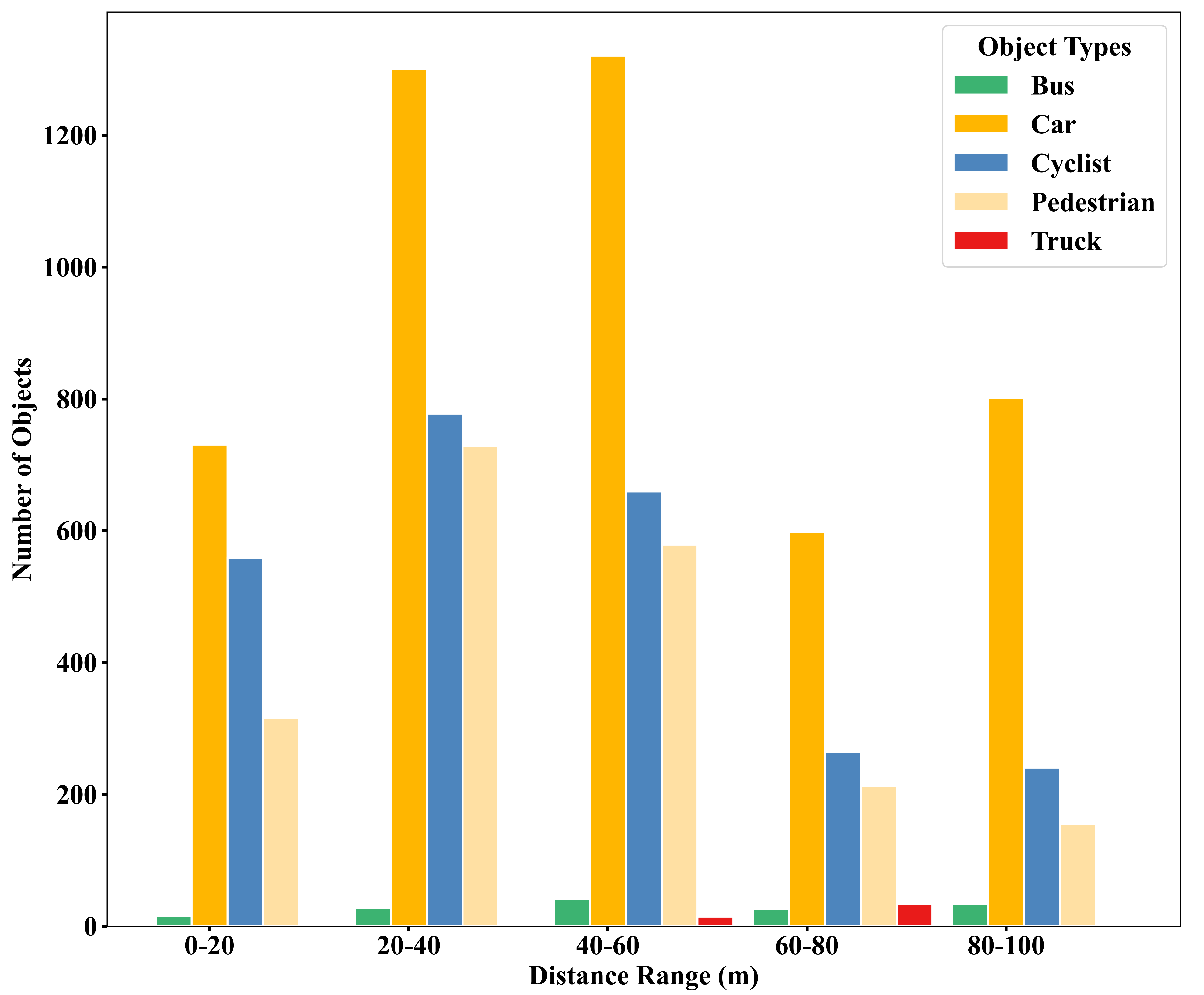}
\label{fig_dis_school}}
\caption{The number of different categories of targets within each twenty-meter distance range. (a) The number of different categories of targets for the urban scenario. (b) The number of different categories of targets for the campus scenario.}
\label{fig_dis}
\end{figure*}

Statistical analyses and visualizations of the annotation results were conducted to reveal the frequency and distribution characteristics of different object categories across various scenarios. Bar charts and pie charts were used to illustrate these distributions. The bar chart (Fig. \ref{fig_dis}) shows the distribution of target objects across different distance ranges in urban and campus scenarios. In urban environments, vehicle categories dominate across all distance bands, particularly within the 40-60 meter range, highlighting the dense vehicle distribution typical of cities. Pedestrian and bicycle counts are concentrated within 20 meters, reflecting common urban mobility patterns. In the campus setting, pedestrians and bicycles constitute a higher proportion of targets, especially in closer proximity, aligning with the low-speed, safety-conscious nature of campus traffic.

\begin{figure*}[!t]
\centering
\subfloat[]{\includegraphics[width=2.5in]{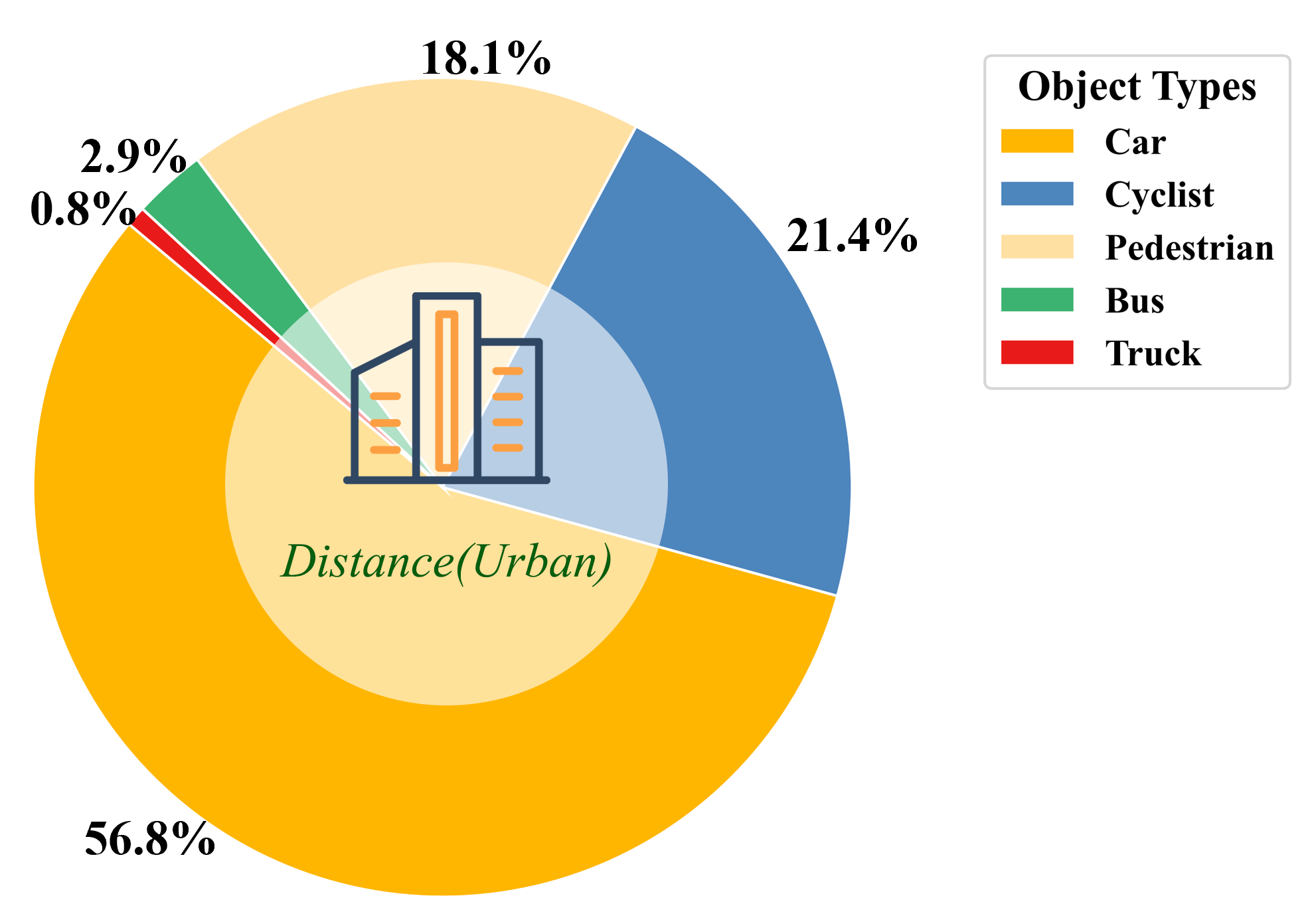}
\label{city}}
\hfil
\subfloat[]{\includegraphics[width=2.5in]{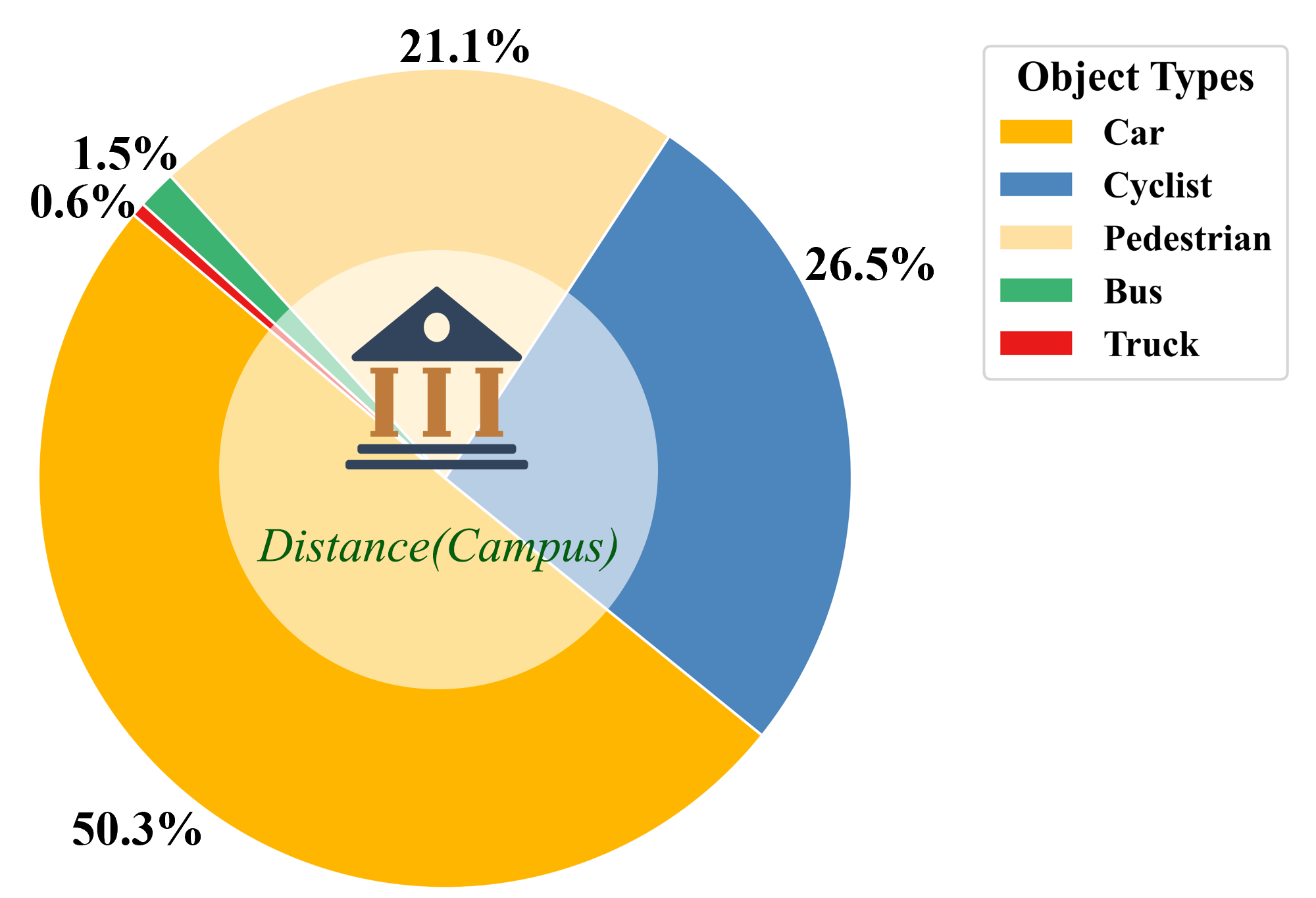}
\label{school}}
\caption{Distribution of different categories of targets. (a) The share of different categories in the urban scenario. (b) The share of different categories in the campus scenario.} 
\label{dis}
\end{figure*}

The pie chart, on the other hand, provides a macro view showing the proportionality of the different object categories in the dataset, which helps to understand the target categories on which the autonomous driving system needs to focus (see Fig. \ref{dis}). This data emphasizes the importance of automated driving systems in recognizing and processing motor vehicles in urban environments, where the vehicle category, especially cars, accounts for more than half of the total (56.8\%). Comparatively, in the campus scenario, pedestrians and cyclists combined accounted for nearly half of the total (21.1\% for pedestrians and 26.5\% for cyclists), a statistic that highlights the high density of foot traffic in the campus environment.

\begin{figure*}[htbp]
    \centering
    \includegraphics[width=0.8\textwidth]{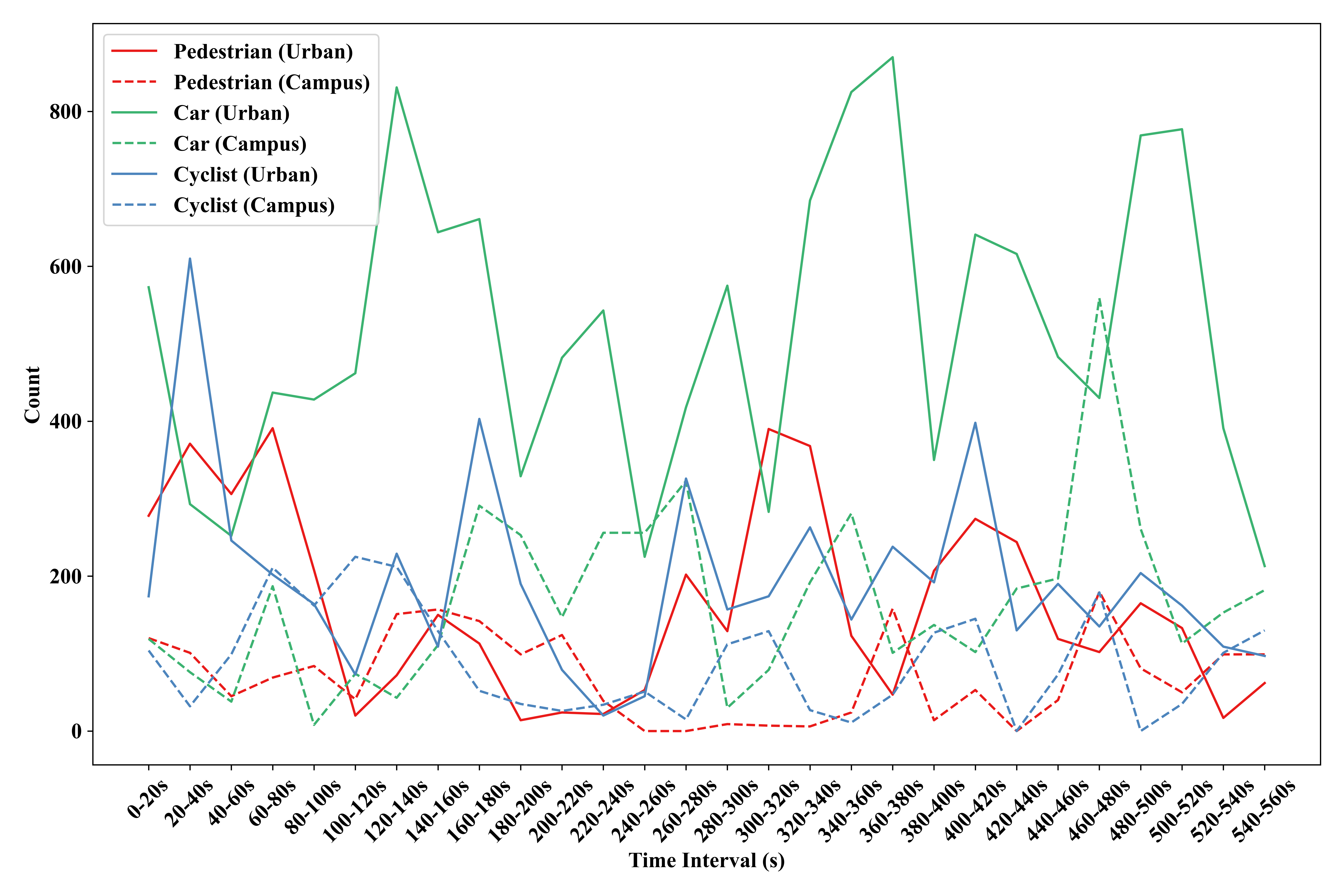}
    \caption{The number of pedestrians, cars, and bicycles changes every 20 seconds in different scenarios, with the solid line representing the urban scene and the dotted line representing the campus scene.}
    \label{fig4}
\end{figure*}

Fig. \ref{fig4} illustrates the count of pedestrians, cars, and cyclists recorded over successive twenty-second intervals in both urban and campus environments, providing a comparative analysis of object distribution. The data shows that urban environments have significantly higher car counts, reflecting dense vehicular traffic, with peaks reaching up to 800 counts. In contrast, campus settings exhibit higher counts of pedestrians and cyclists, aligning with common campus activities such as walking and bicycling. Specifically, the number of pedestrians and cyclists in campus scenarios frequently surpasses 200 counts, indicating a predominance of non-vehicular movement in these areas.

\begin{figure*}[!h]
\centering
\begin{minipage}[b]{0.36\linewidth}
	\subfloat[]{\label{z:(a)}
	\includegraphics[width=\linewidth]{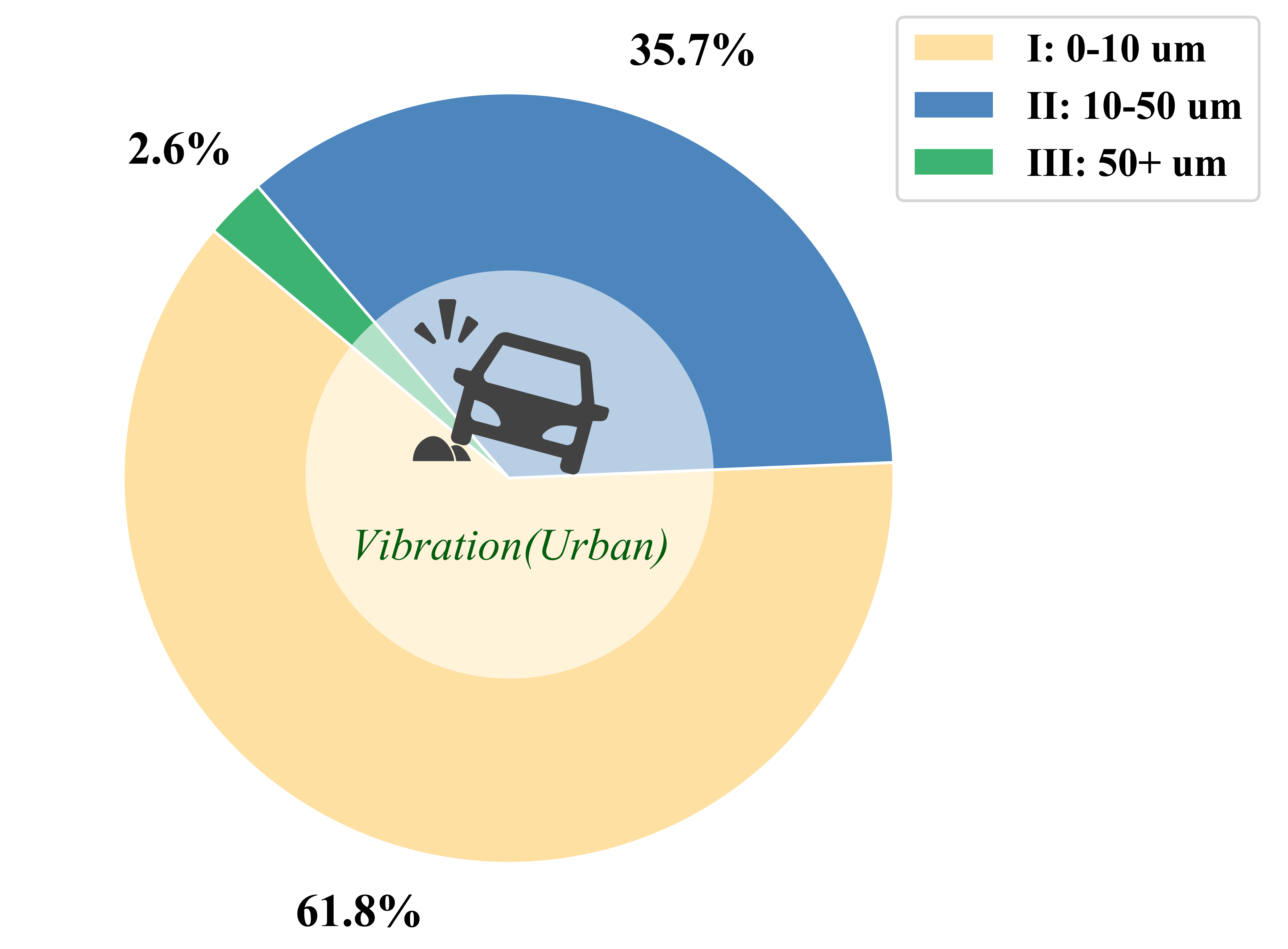}}
        \vspace{0.2cm}
\end{minipage}
\hspace{0.02\linewidth}
\begin{minipage}[b]{0.2\linewidth}
	\subfloat[]{\label{z:(b)}
	\includegraphics[width=\linewidth]{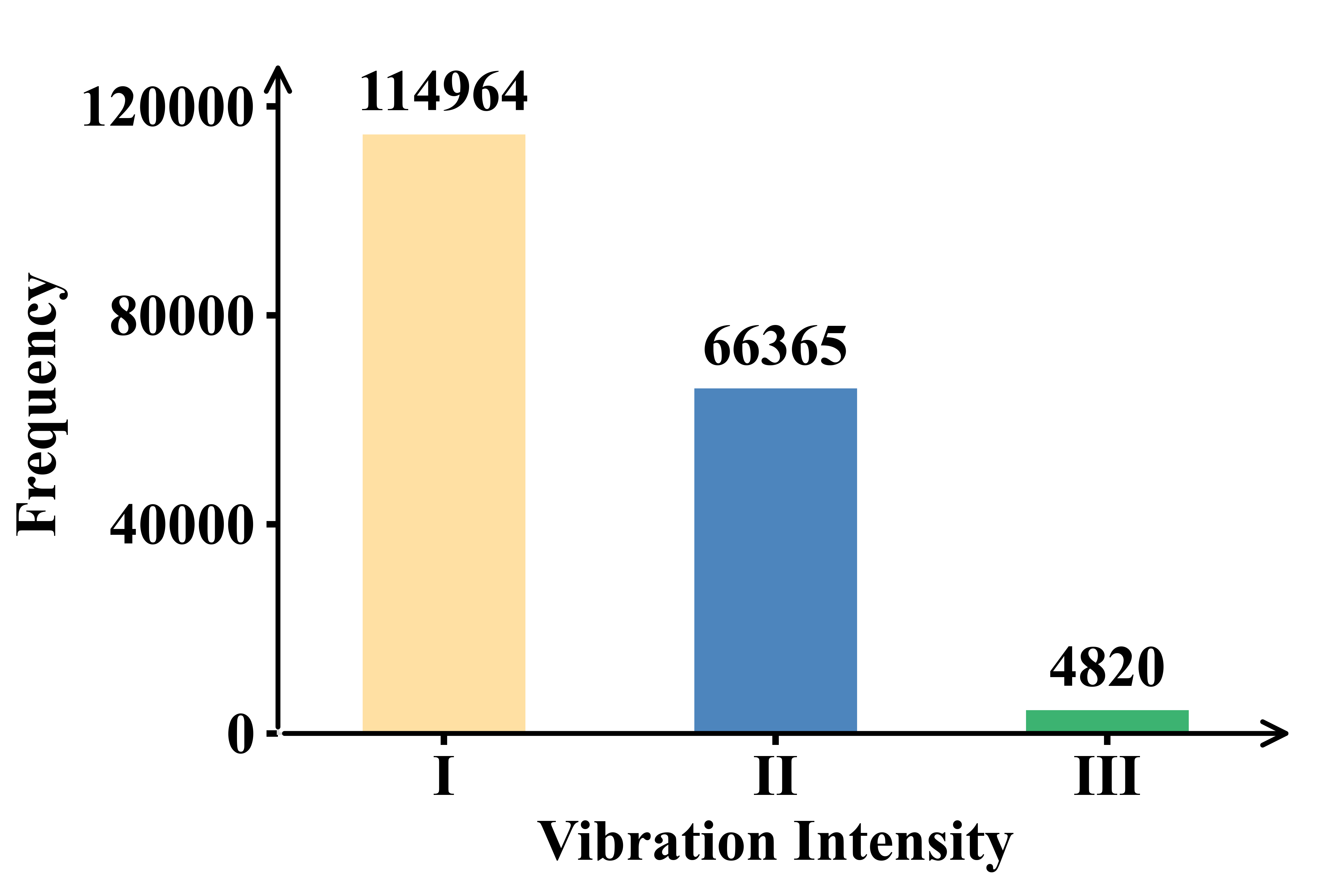}} \\ 
	\subfloat[]{\label{z:(c)}
	\includegraphics[width=\linewidth]{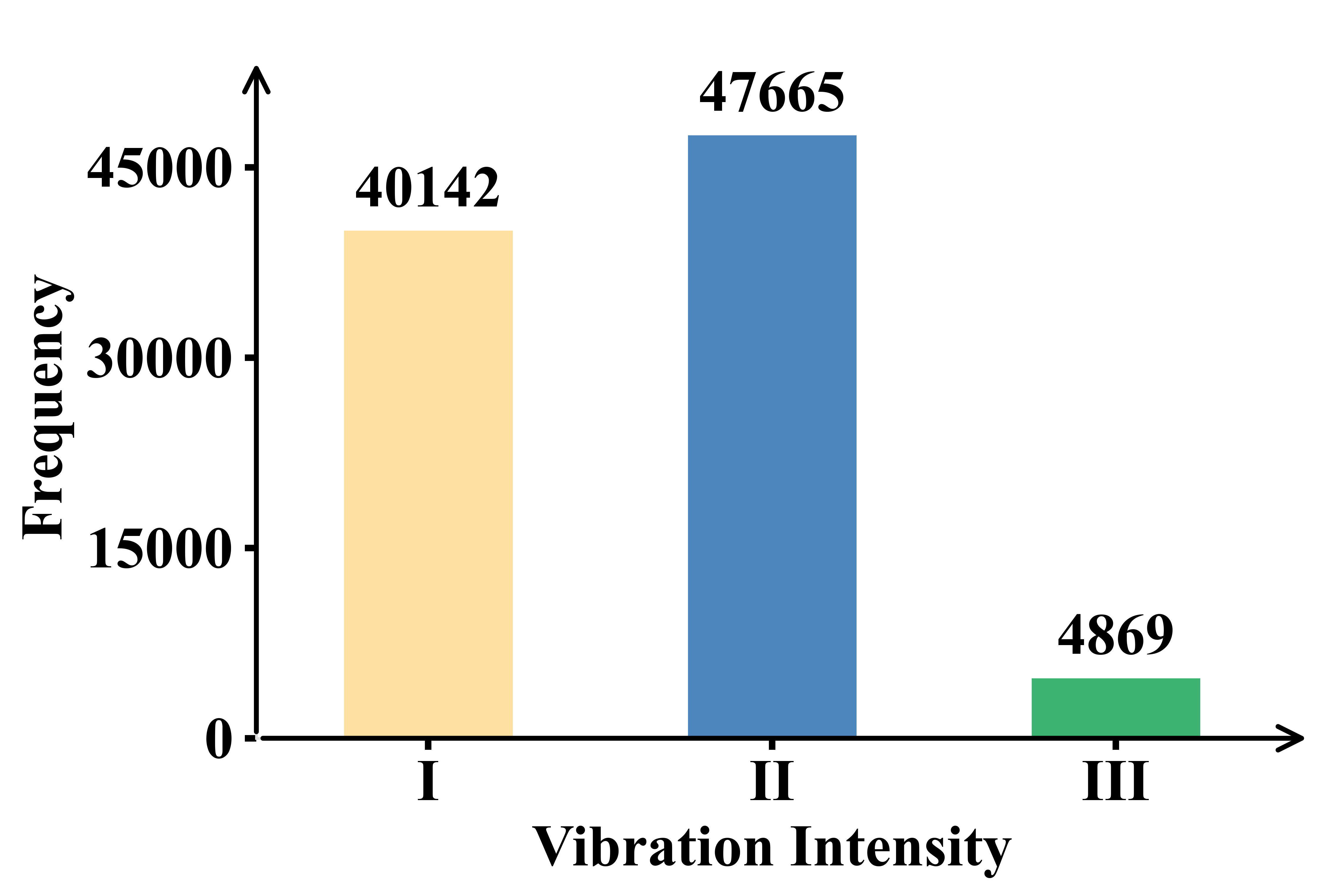}}
\end{minipage} 
\hspace{0.02\linewidth}
\begin{minipage}[b]{0.36\linewidth}
	\subfloat[]{\label{z:(d)}
	\includegraphics[width=\linewidth]{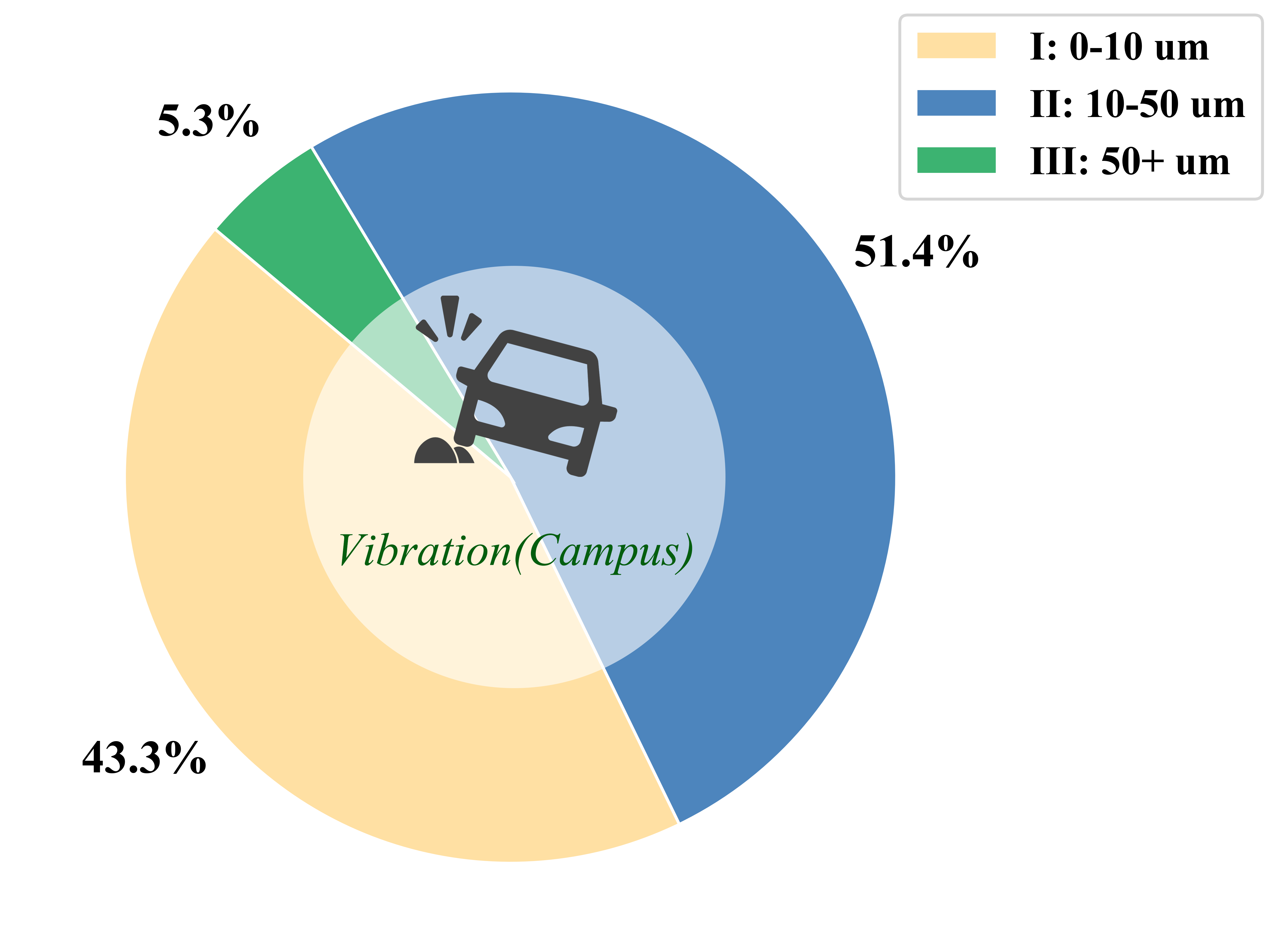}}
 \vspace{0.33cm}
\end{minipage}

\caption{Comparison of vibration data in urban and campus, where 0-10 $\upmu$m is classified as Class I, 10-50 $\upmu$m as Class II, and 50+ $\upmu$m as Class III. (a) Vibration distribution (Urban); (b) Vibration intensity (Urban); (c) Vibration intensity (Campus); (d) Vibration distribution (Campus). }
\label{z}

\end{figure*}

In addition, the analysis of vibration and light intensity distributions provides us with valuable information on how environmental conditions specifically affect the autonomous driving sensing system. In the case of the vibration sensor, the gyroscope acceleration was measured, and integral operations were performed to find the amplitude of vibration in the Z-axis direction. The collected data were then categorized into three levels based on the amplitude values in different cases. Given that the vibration displacement of the Z-axis when the vehicle passes through a speed bump is about 100 $\upmu$m, while the vibration in the normal state stays below 10 $\upmu$m, the vibration intensity is classified into three classes: 0-10 $\upmu$m as class I, 10-50 $\upmu$m as class II, and more than 50 $\upmu$m as class III. The specific cases of passing through a speed bump and crossing a pit with the left front wheel were marked in the corresponding sound data. The specific distribution is shown in Fig. \ref{z}. The data indicates that in urban environments, a significant majority of vibrations fall into class I (61.8\%), with smaller proportions in class II (35.7\%) and class III (2.6\%). Conversely, in campus environments, class I vibrations also dominate (51.4\%), but with relatively higher proportions of class II (43.3\%) and class III (5.3\%) vibrations, suggesting more varied and frequent minor obstacles. This aligns with the characteristic presence of numerous speed bumps in campus environments.

The light sensor uses a high-sensitivity photodetector, with a high-precision linear amplification circuit to convert the change in luminous flux per unit area into a change in current intensity, and the real-time light intensity by conversion, the specific data conversion formula as shown in Equation \ref{deqn_ex1}, where E is the intensity of the light, the unit of Lux, and A is the value of the current captured by the collector, the unit of mA.

\begin{equation}
\label{deqn_ex1}
E = (1.25 \cdot A - 5) \cdot 10^4
\end{equation}

\begin{figure*}[!h]
\centering
\begin{minipage}[b]{0.36\linewidth}
	\subfloat[]{\label{z:(a)}
	\includegraphics[width=\linewidth]{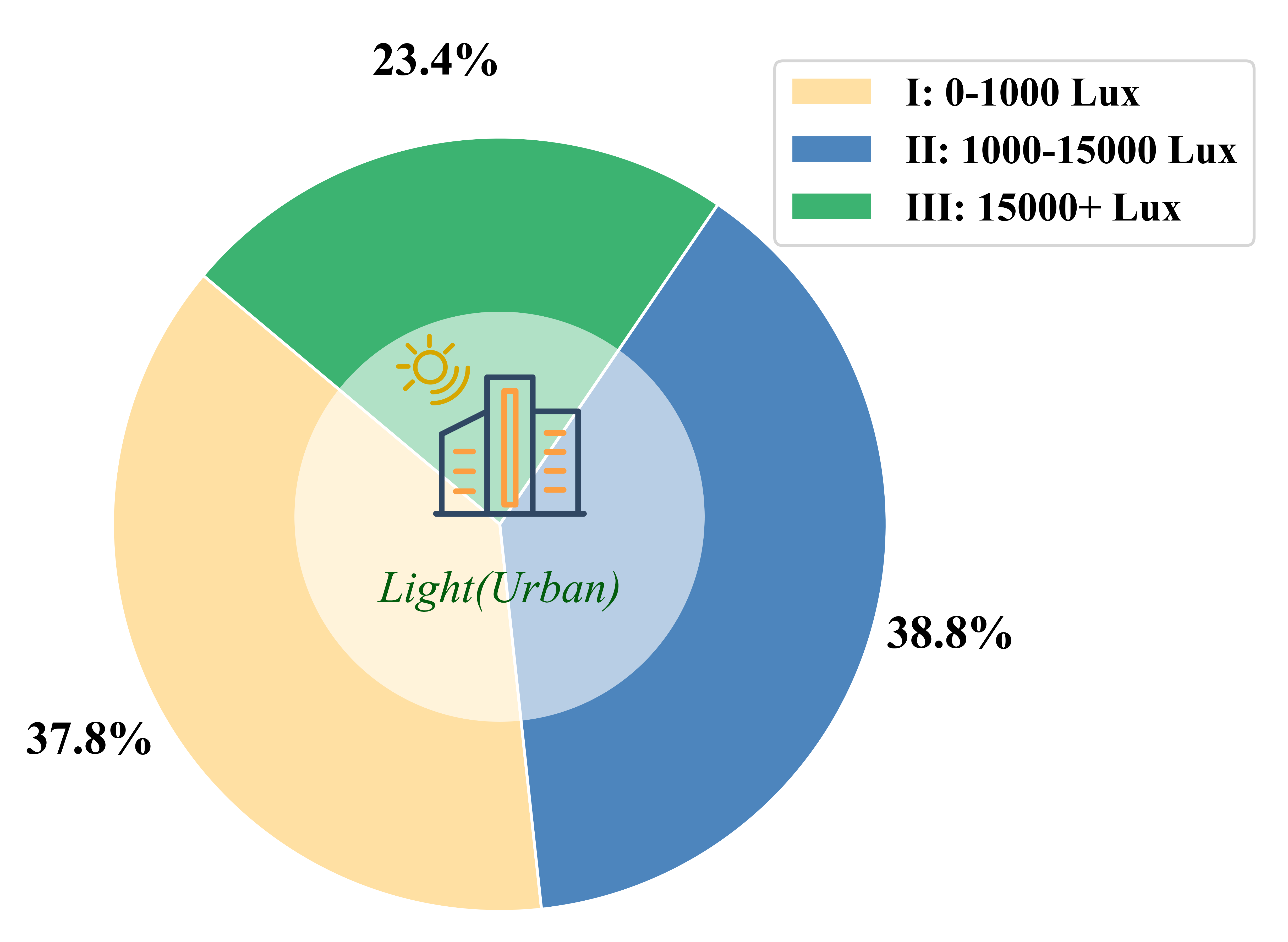}}
        \vspace{0.2cm}
\end{minipage}
\hspace{0.02\linewidth}
\begin{minipage}[b]{0.2\linewidth}
	\subfloat[]{\label{z:(b)}
	\includegraphics[width=\linewidth]{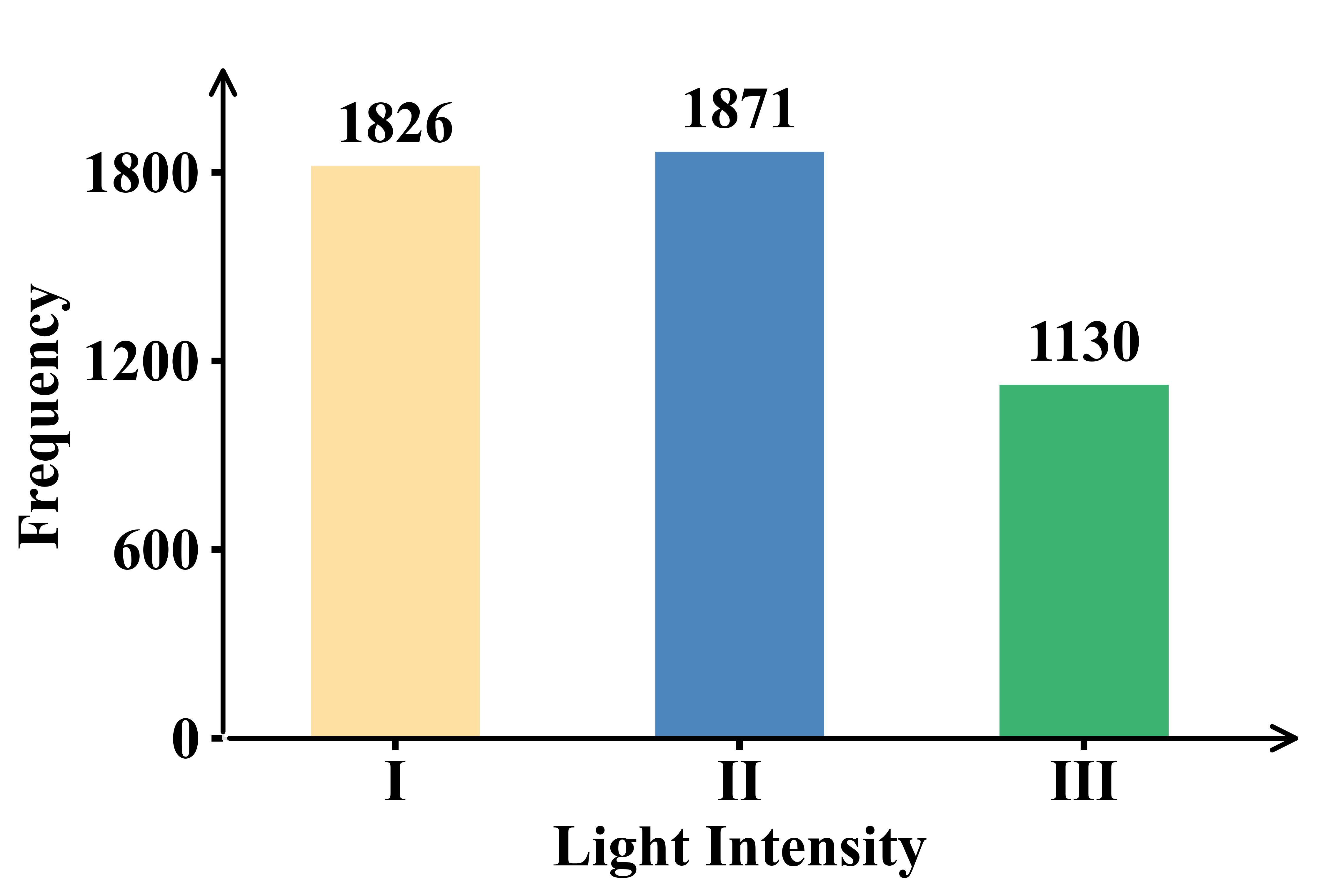}} \\ 
	\subfloat[]{\label{z:(c)}
	\includegraphics[width=\linewidth]{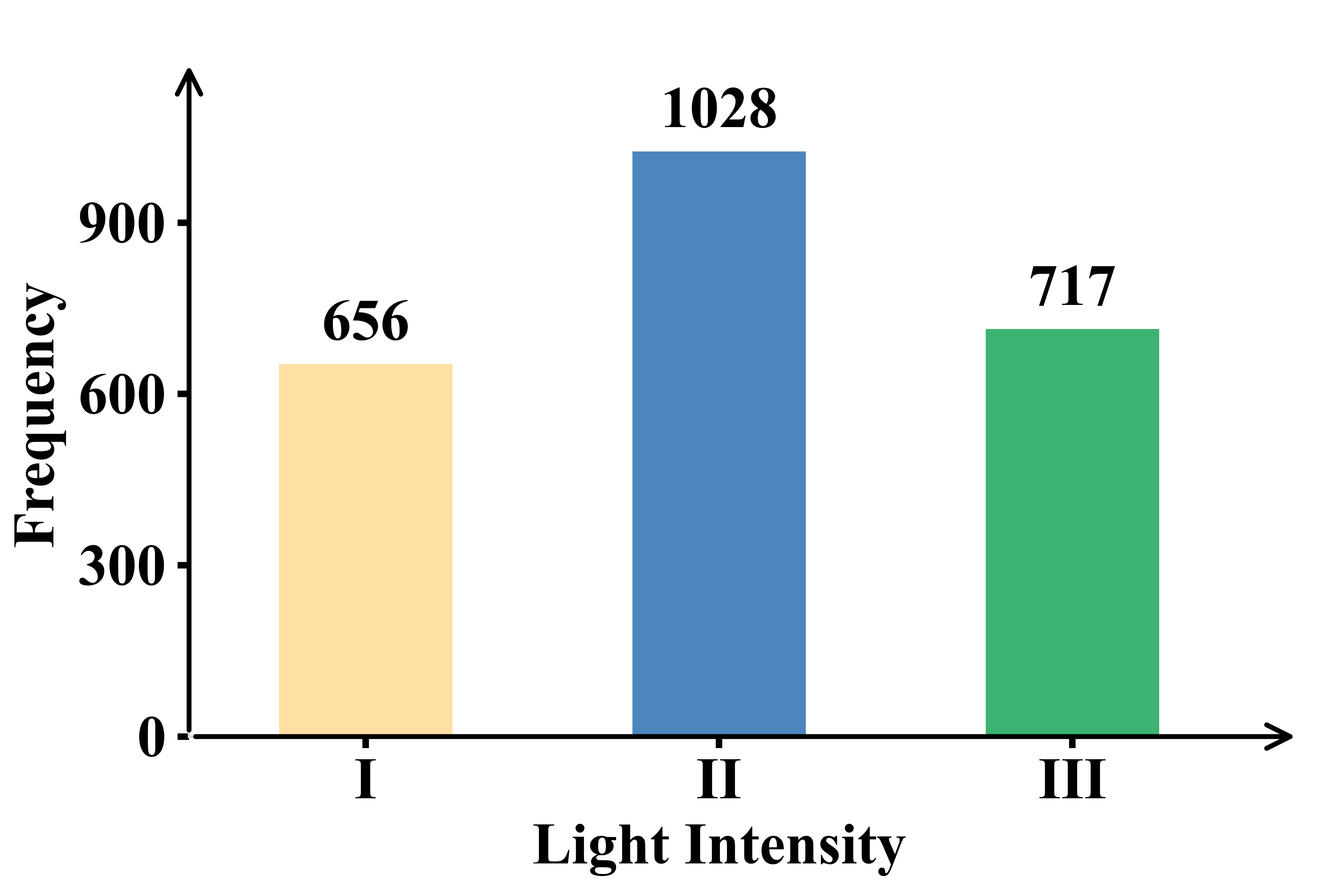}}
\end{minipage} 
\hspace{0.02\linewidth}
\begin{minipage}[b]{0.36\linewidth}
	\subfloat[]{\label{z:(d)}
	\includegraphics[width=\linewidth]{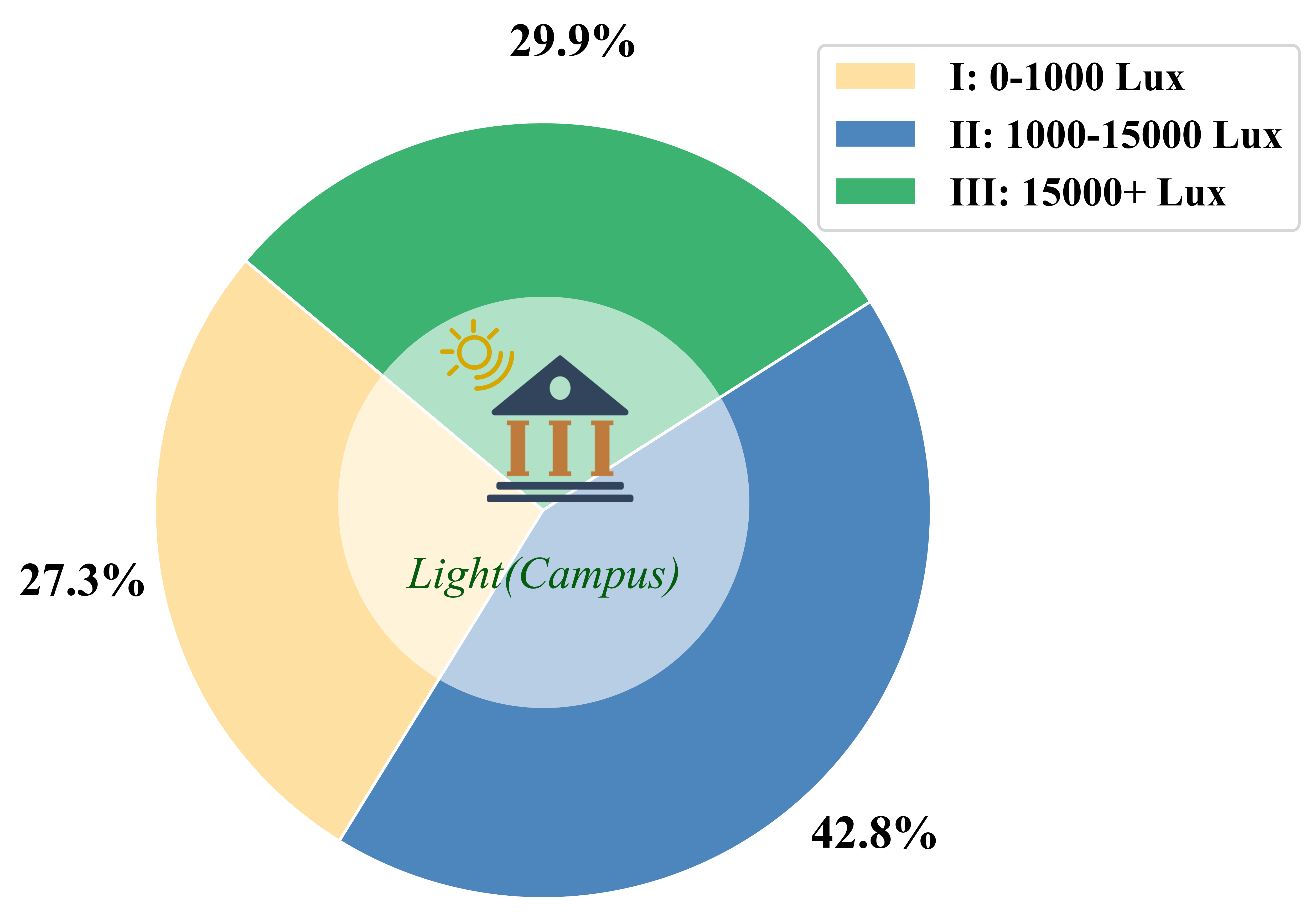}}
 \vspace{0.33cm}
\end{minipage}

\caption{Comparison of lighting intensity in urban and campus, where 0-1000 Lux is Class I, 1000-15000 Lux is Class II, and above 15000 Lux is Class III. (a) Light distribution (Urban); (b) Light intensity (Urban); (c) Light intensity (Campus); (d) Light distribution (Campus). }
\label{l}

\end{figure*}

Given that light intensity at midday can reach up to about 70,000 Lux and light intensity at night is usually below 1,000 Lux, light intensity is categorized into three levels: 0-1,000 Lux, 1,000-15,000 Lux, and above 15,000 Lux. The corresponding acoustic data are annotated with data for scenario-varying situations, such as passing under a bridge or a building. This hierarchical approach not only captures the light variations at different periods but also reflects the perceptual challenges that an autonomous driving system may encounter when passing through different locations. Fig. \ref{l} visualizes the details of these distributions. In urban environments, the distribution shows that 38.8\% of light intensity measurements fall into the 1,000-15,000 Lux category (Class II), followed by 37.8\% in the 0-1,000 Lux range (Class I), and 23.4\% above 15,000 Lux (Class III). Conversely, in campus environments, 42.8\% of measurements fall into Class II, 27.3\% into Class I, and 29.9\% into Class III, indicating more consistent moderate lighting conditions typically found in such settings.

Accurate perception of the surrounding environment is crucial in the research and development of autonomous driving systems. Target detection, as a core component of the perception system, plays a decisive role in ensuring road traffic safety and providing a smooth autonomous driving experience. Therefore, target detection is taken as an example to explore the impact of different types of data sensing on its performance.

Statistical analysis of the vibration data, highlights the stability challenges encountered by vehicles passing through speed bumps and potholes, which have a direct impact on the performance of target detection algorithms. Specifically, when vehicles experience high levels of vibration, the image and point cloud data captured by sensors - particularly cameras and LIDAR - may be disturbed, compromising the accuracy and reliability of target detection. Therefore, vibration sensor data becomes an important reference for adapting and evaluating algorithms to real-world road conditions.

Similarly, the impact of changes in light intensity on the sensing system should not be ignored. Autonomous driving systems need to maintain stable target detection performance under different lighting conditions, from dawn to dusk, as well as at night and in various weather conditions. Data from light sensors, especially in high dynamic range environments, provide intuitive insights into the impact of different lighting conditions on the performance of detection algorithms. For example, in bright light or backlight conditions, the sensing system may have difficulty recognizing the target in front of it; in low-light environments, dark or low-reflectivity objects may be missed. By simulating and analyzing these challenges, the algorithms were optimized to ensure accurate detection and classification of obstacles on the road under all conditions.

\subsection{Data Visualization}
\begin{figure*}[!ht]
    \centering
    \hspace{1cm} Camera \hspace{4.5cm} LiDAR \hspace{4.5cm} 4D Radar 
    \vskip\baselineskip
    \begin{minipage}[b]{0.05\textwidth}
        \footnotesize(a)
        \vspace{1.2cm}
    \end{minipage}
     \begin{minipage}[b]{0.3\textwidth}        
        \includegraphics[width=\textwidth,height=3cm]{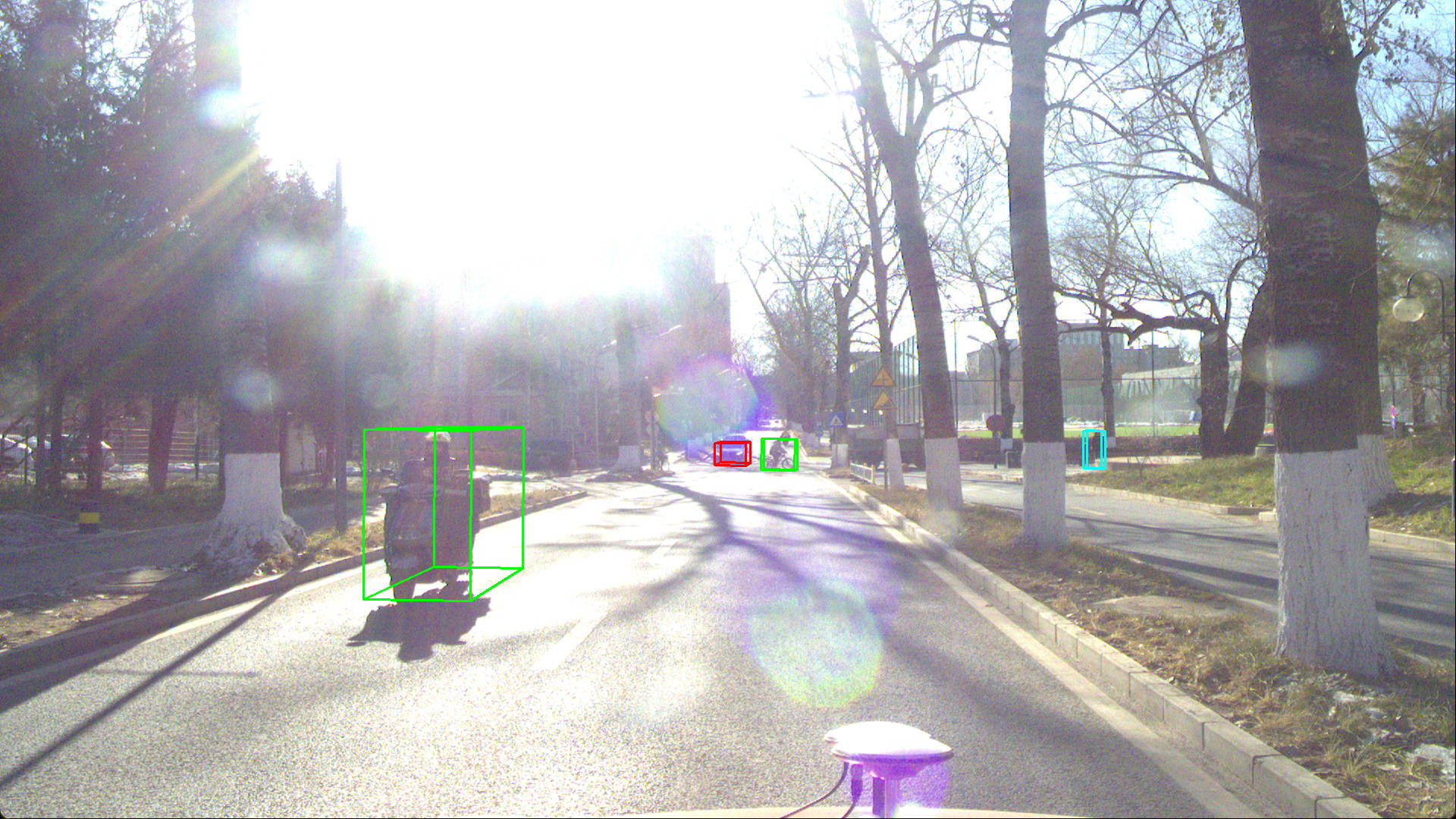}
    \end{minipage}
    \hfill
    \begin{minipage}[b]{0.3\textwidth}
        \includegraphics[width=\textwidth,height=3cm]{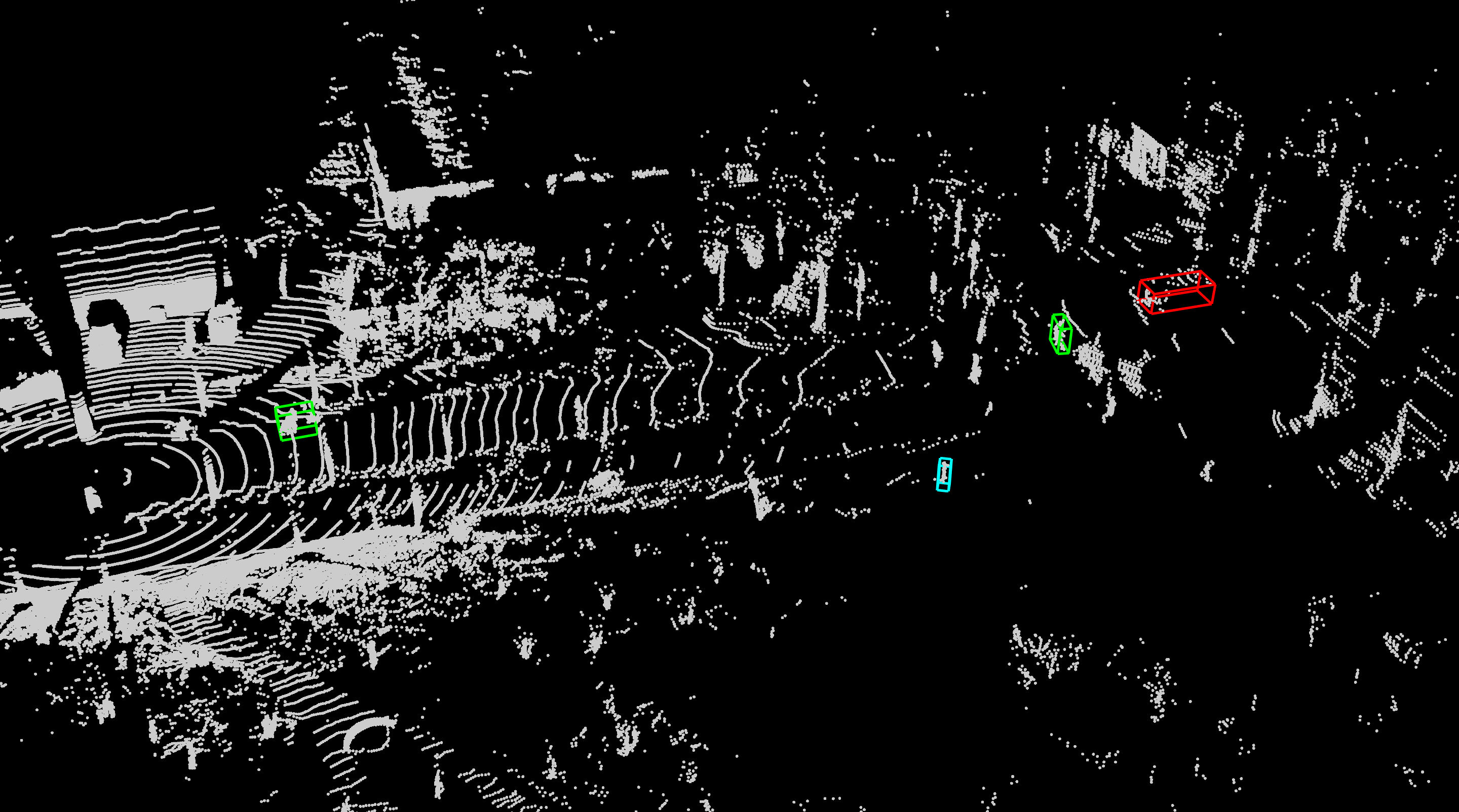}
    \end{minipage}
    \hfill
    \begin{minipage}[b]{0.3\textwidth}
        \includegraphics[width=\textwidth,height=3cm]{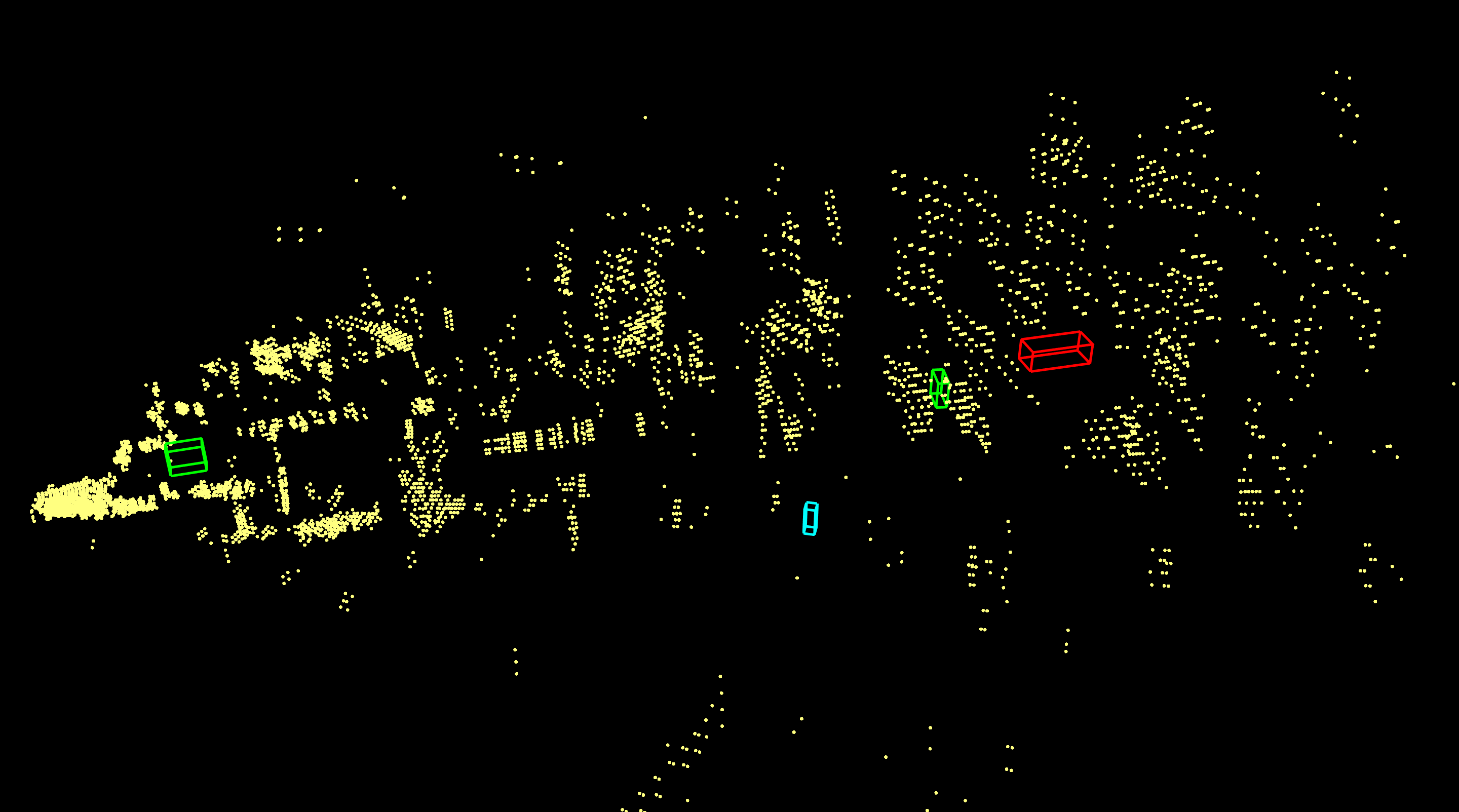}
    \end{minipage}
    
    \vskip\baselineskip
    \begin{minipage}[b]{0.05\textwidth}
        \footnotesize(b)
        \vspace{1.2cm}
    \end{minipage}
    \begin{minipage}[b]{0.3\textwidth}
        \includegraphics[width=\textwidth,height=3cm]{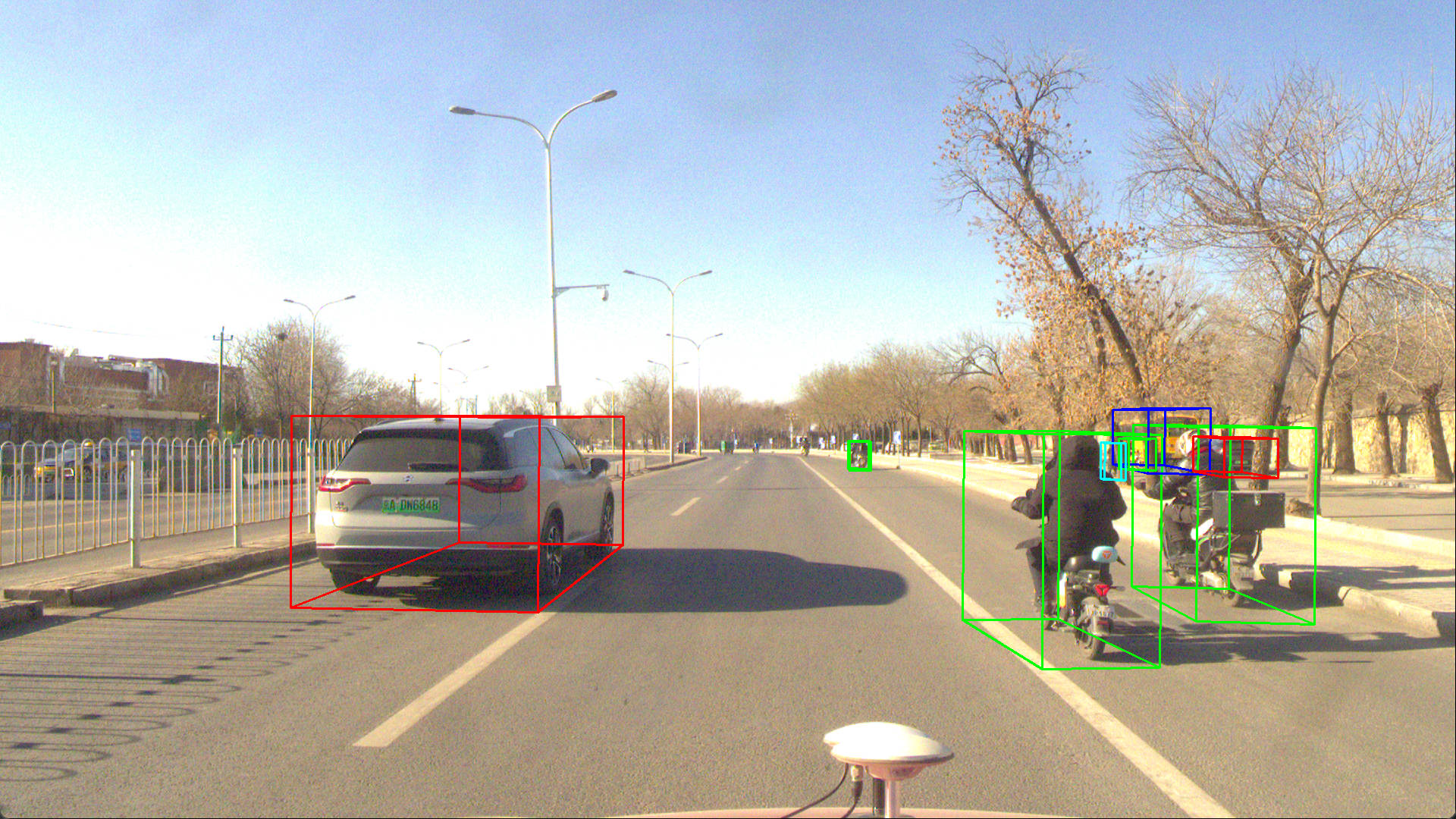}
    \end{minipage}
    \hfill
    \begin{minipage}[b]{0.3\textwidth}
        \includegraphics[width=\textwidth,height=3cm]{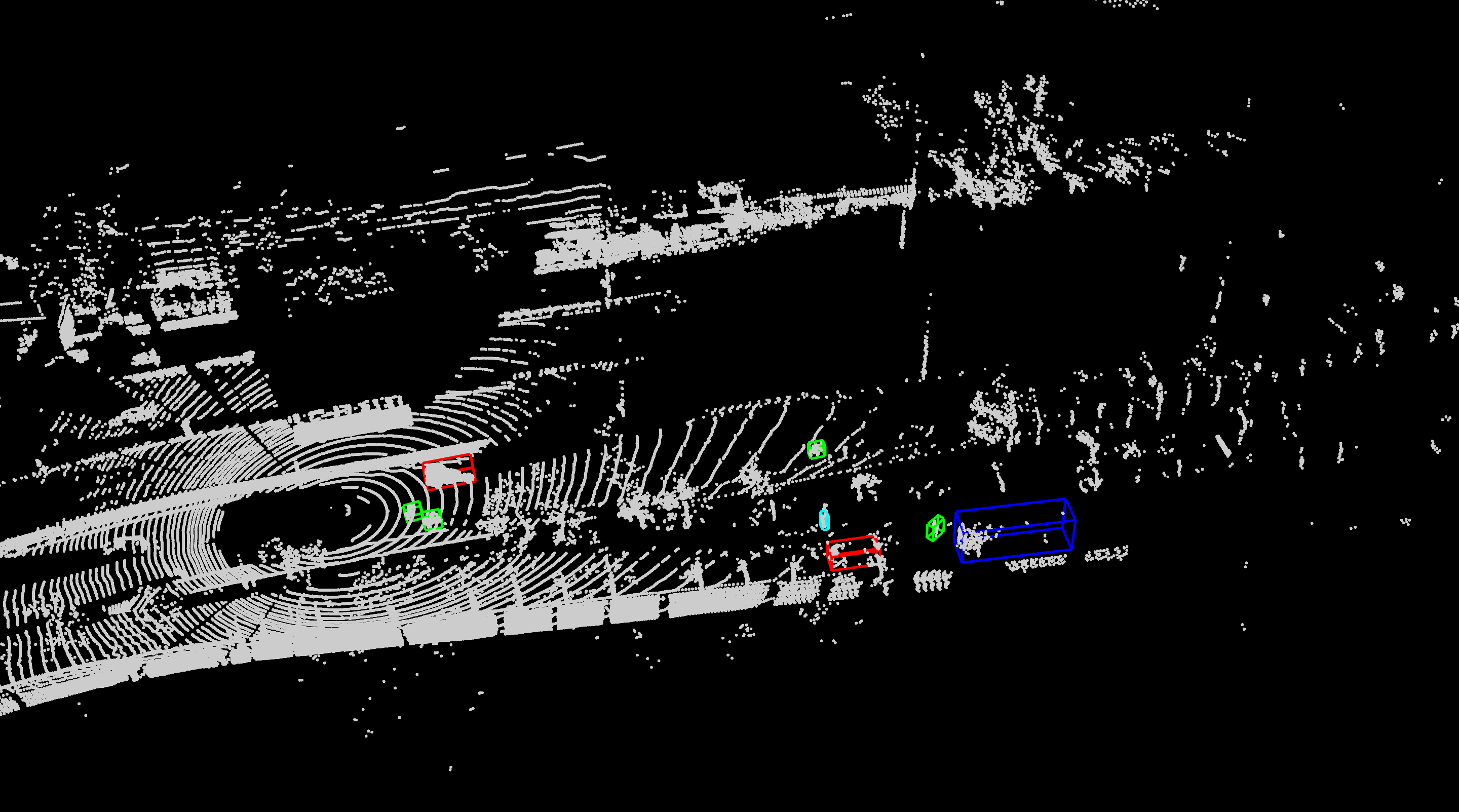}
    \end{minipage}
    \hfill
    \begin{minipage}[b]{0.3\textwidth}
        \includegraphics[width=\textwidth,height=3cm]{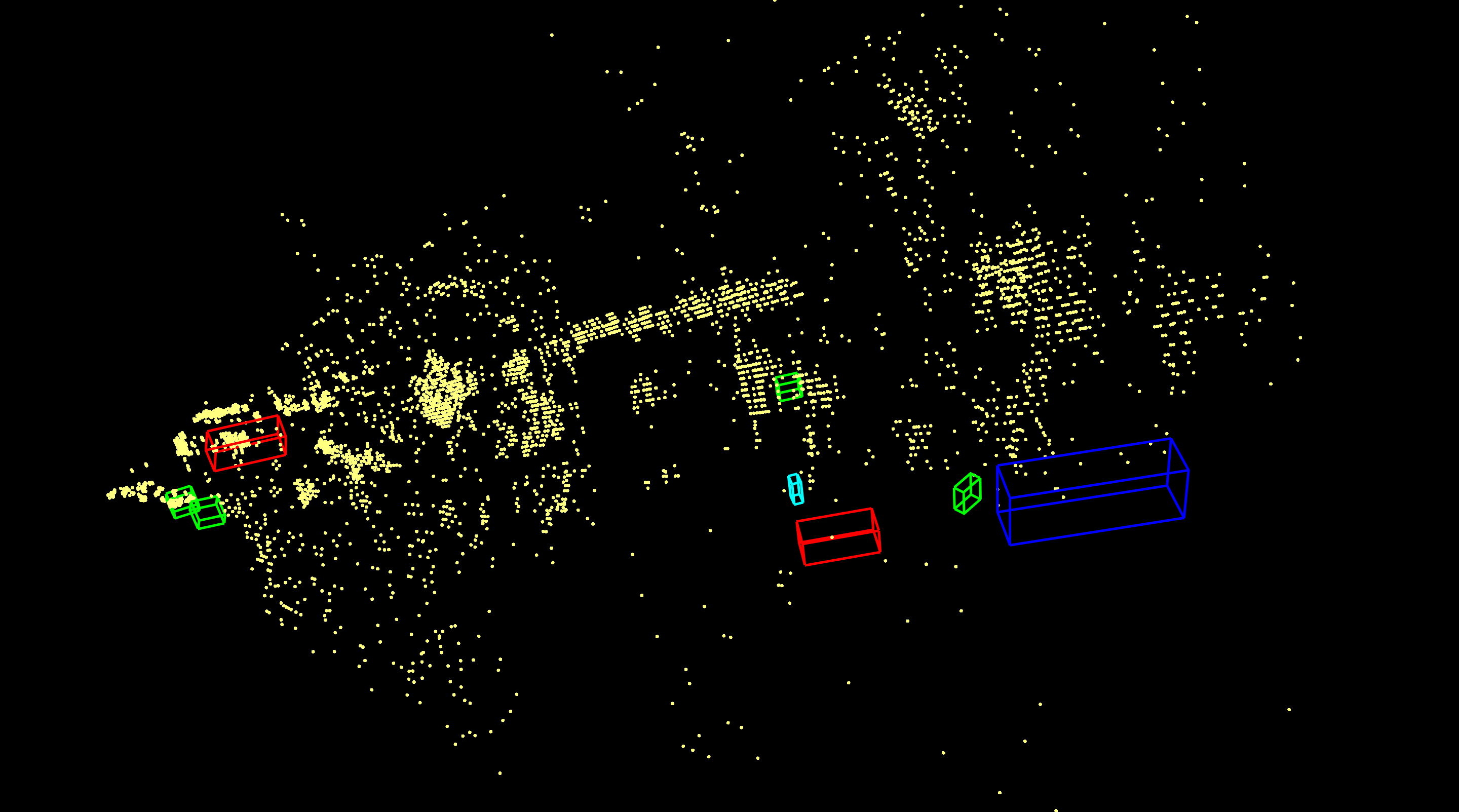}
    \end{minipage}

    \vskip\baselineskip
    \begin{minipage}[b]{0.05\textwidth}
        \footnotesize(c)
        \vspace{1.2cm}
    \end{minipage}
    \begin{minipage}[b]{0.3\textwidth}
        \includegraphics[width=\textwidth,height=3cm]{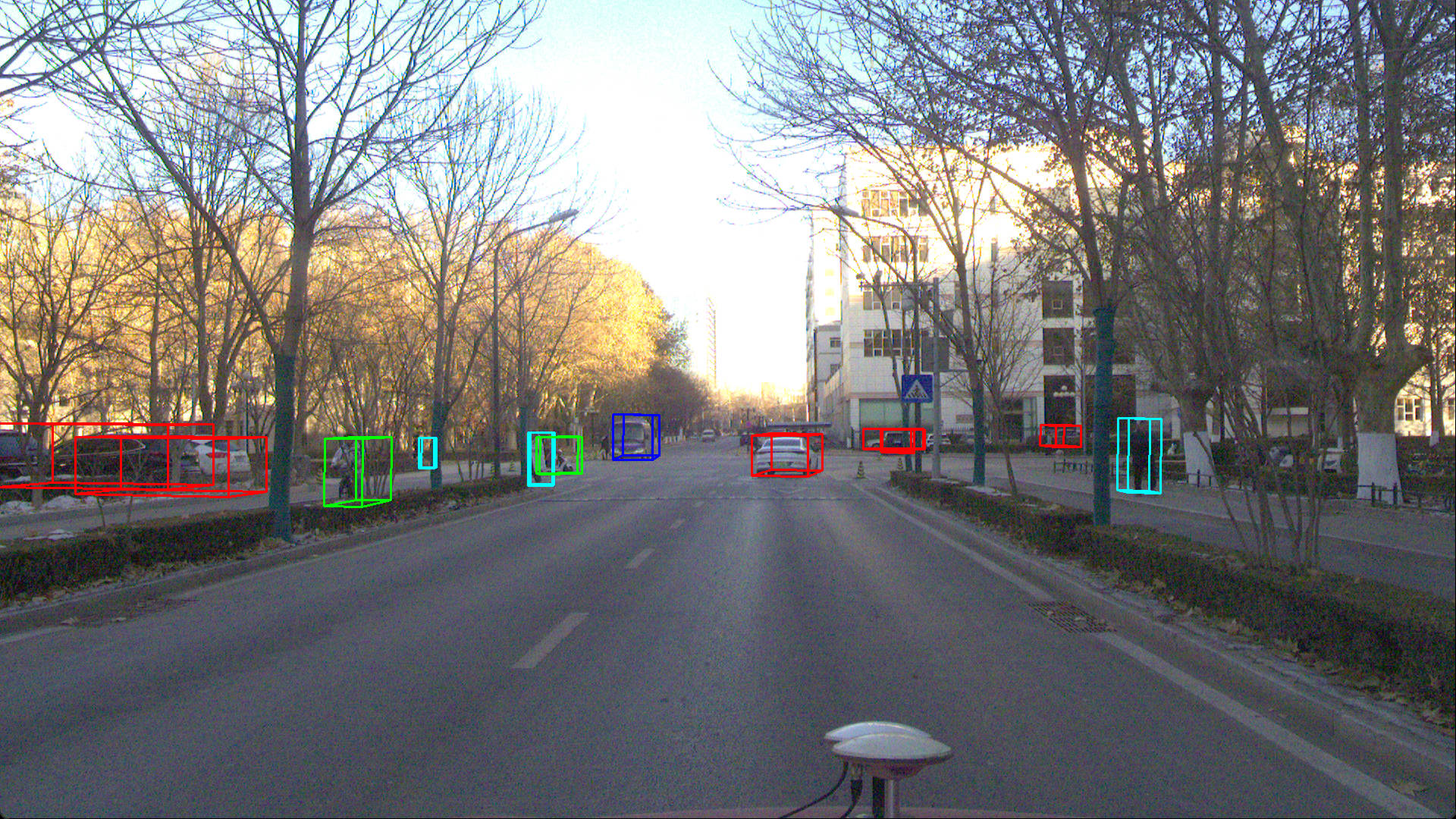}
    \end{minipage}
    \hfill
    \begin{minipage}[b]{0.3\textwidth}
        \includegraphics[width=\textwidth,height=3cm]{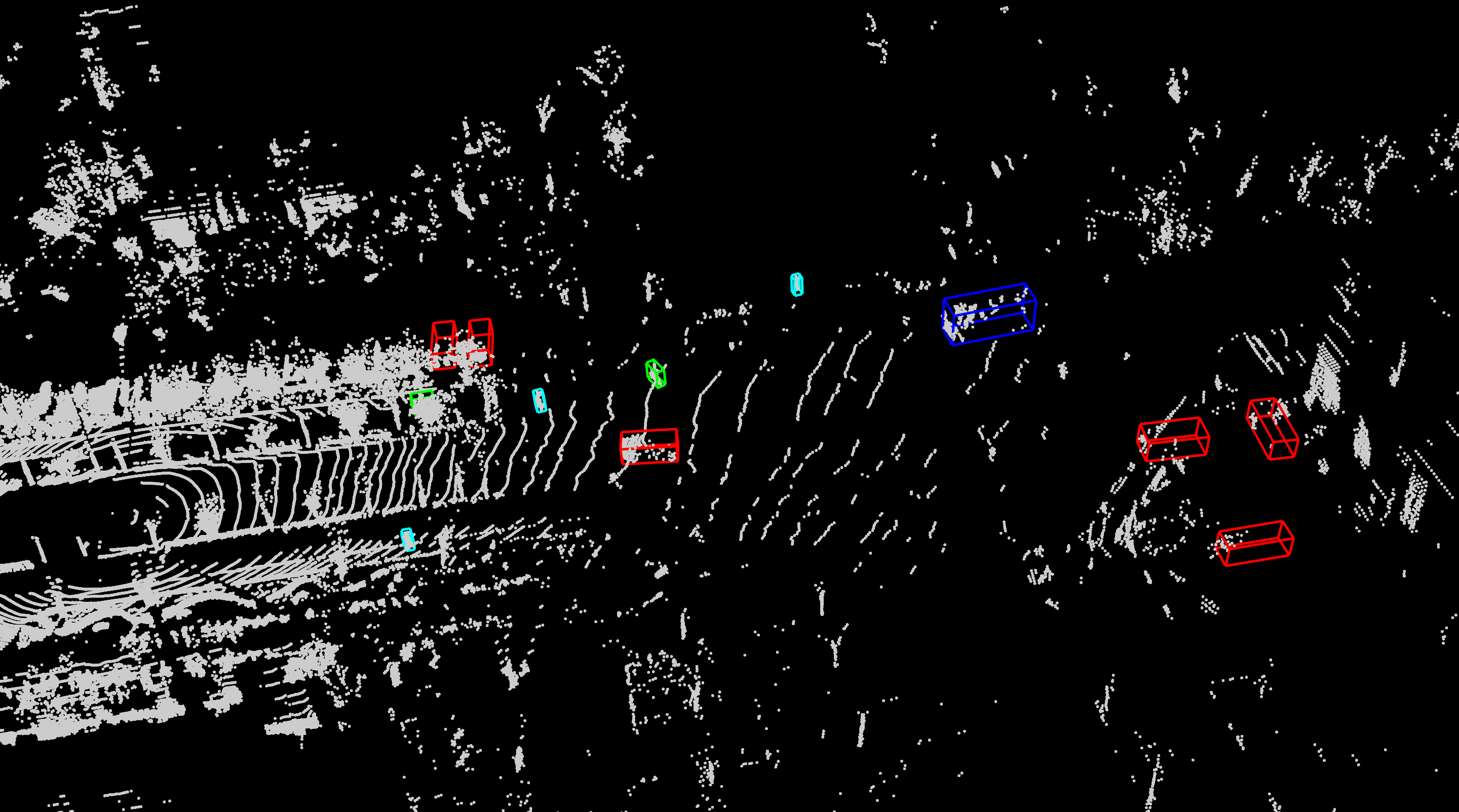}
    \end{minipage}
    \hfill
    \begin{minipage}[b]{0.3\textwidth}
        \includegraphics[width=\textwidth,height=3cm]{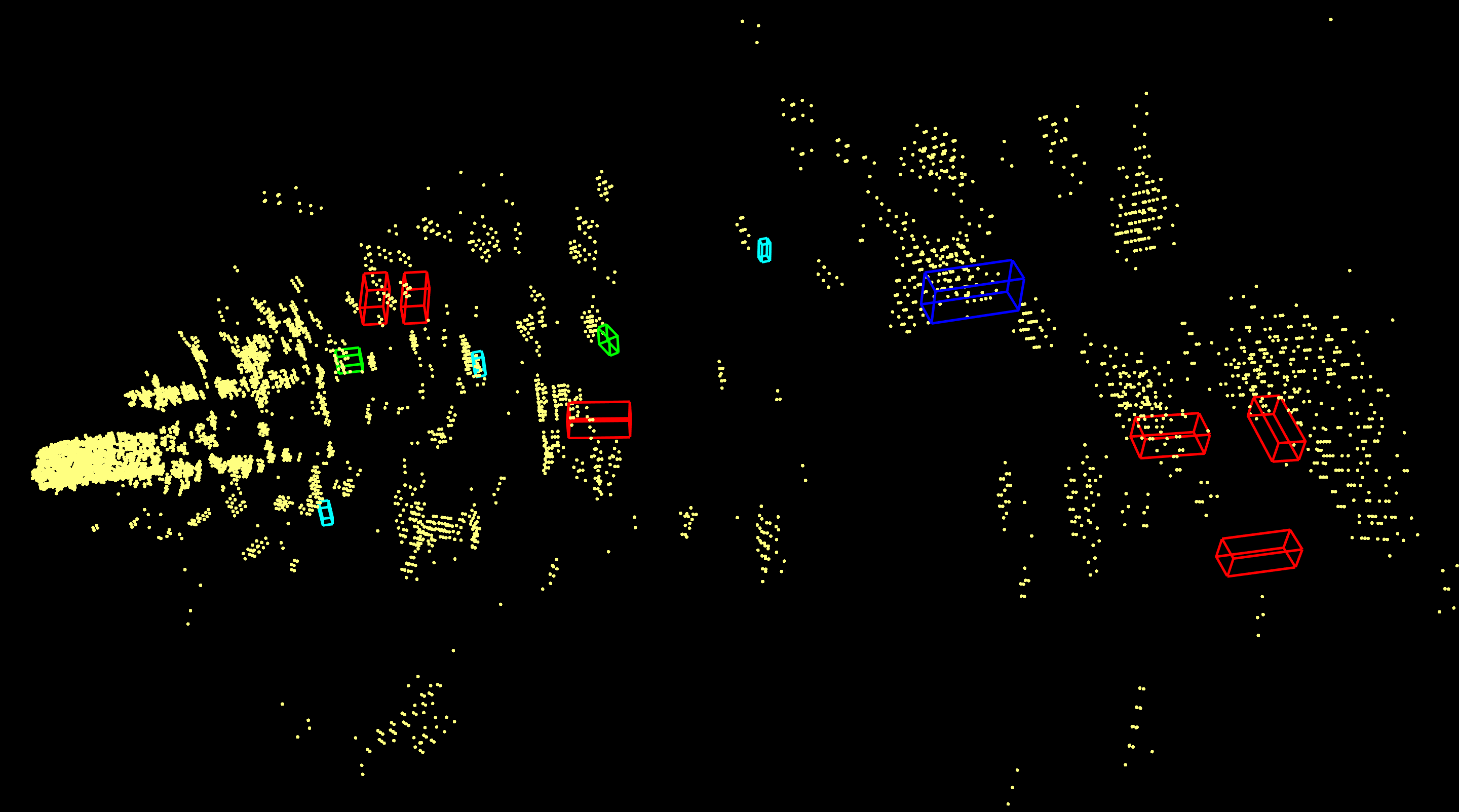}
    \end{minipage}

    \vskip\baselineskip
    \begin{minipage}[b]{0.05\textwidth}
        \footnotesize(d)
        \vspace{1.2cm}
    \end{minipage}
    \begin{minipage}[b]{0.3\textwidth}
        \includegraphics[width=\textwidth,height=3cm]{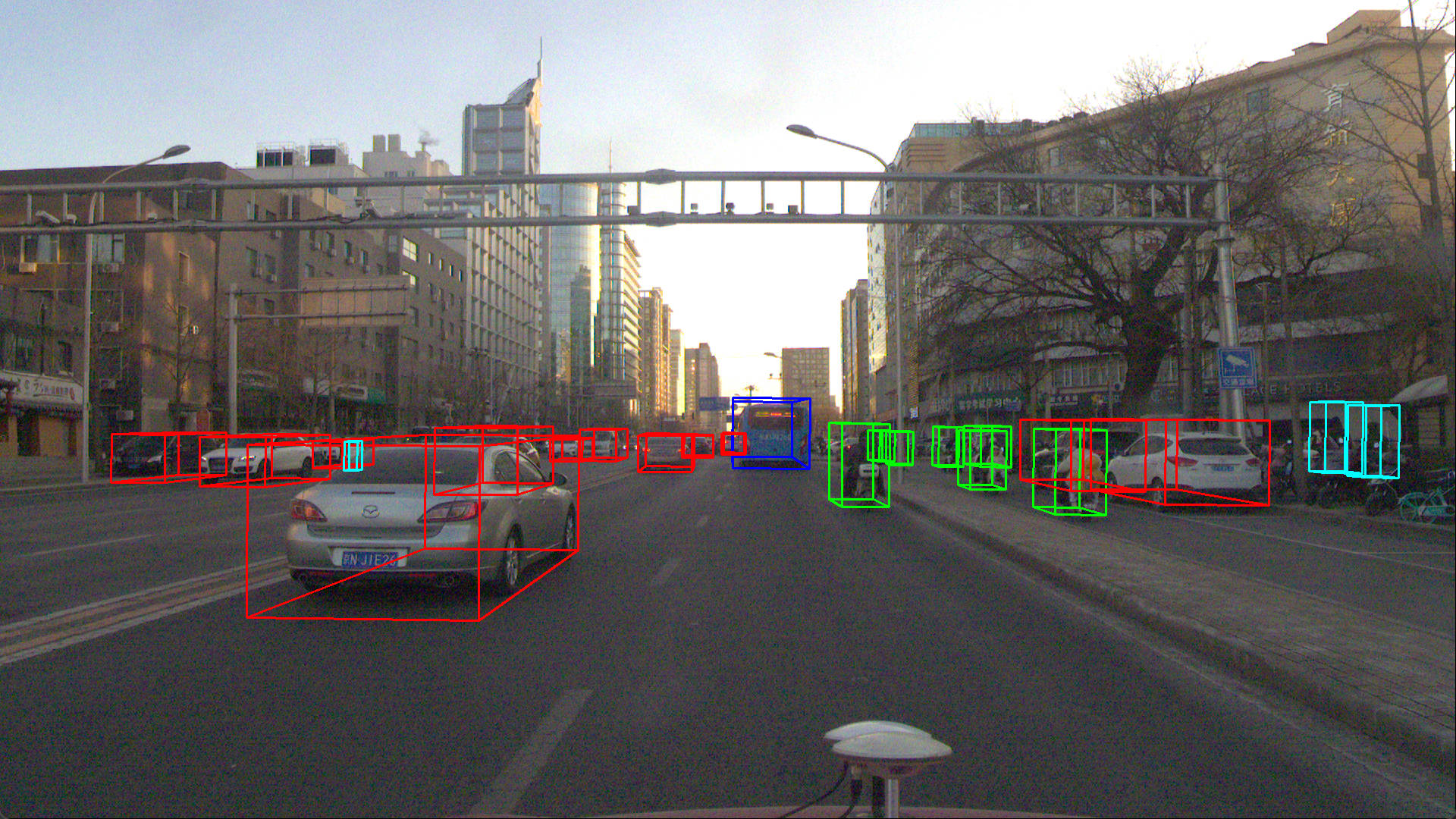}
    \end{minipage}
    \hfill
    \begin{minipage}[b]{0.3\textwidth}
        \includegraphics[width=\textwidth,height=3cm]{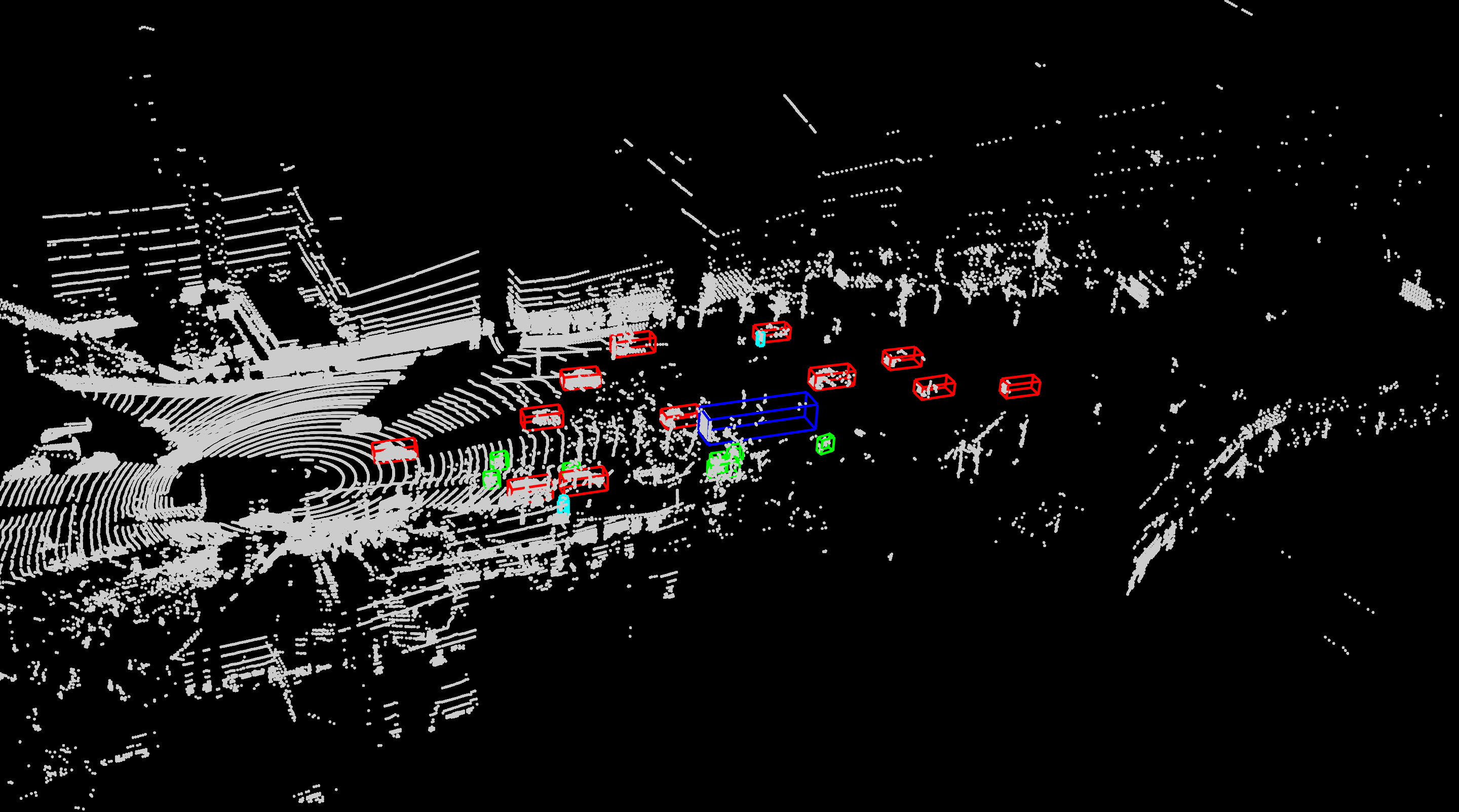}
    \end{minipage}
    \hfill
    \begin{minipage}[b]{0.3\textwidth}
        \includegraphics[width=\textwidth,height=3cm]{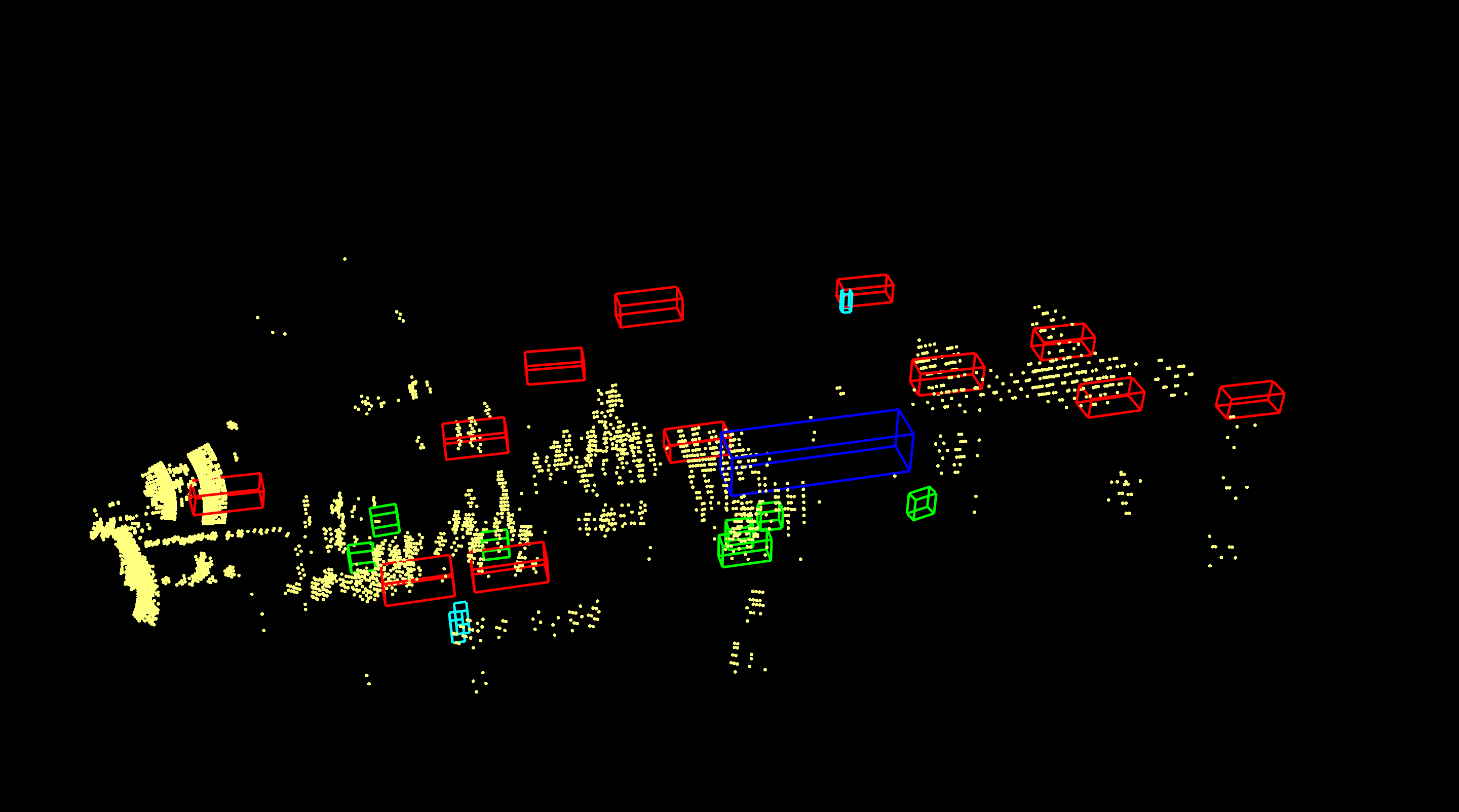}
    \end{minipage}

    \vskip\baselineskip
    \begin{minipage}[b]{0.05\textwidth}
        \footnotesize(e)
        \vspace{1.2cm}
    \end{minipage}
    \begin{minipage}[b]{0.3\textwidth}
        \includegraphics[width=\textwidth,height=3cm]{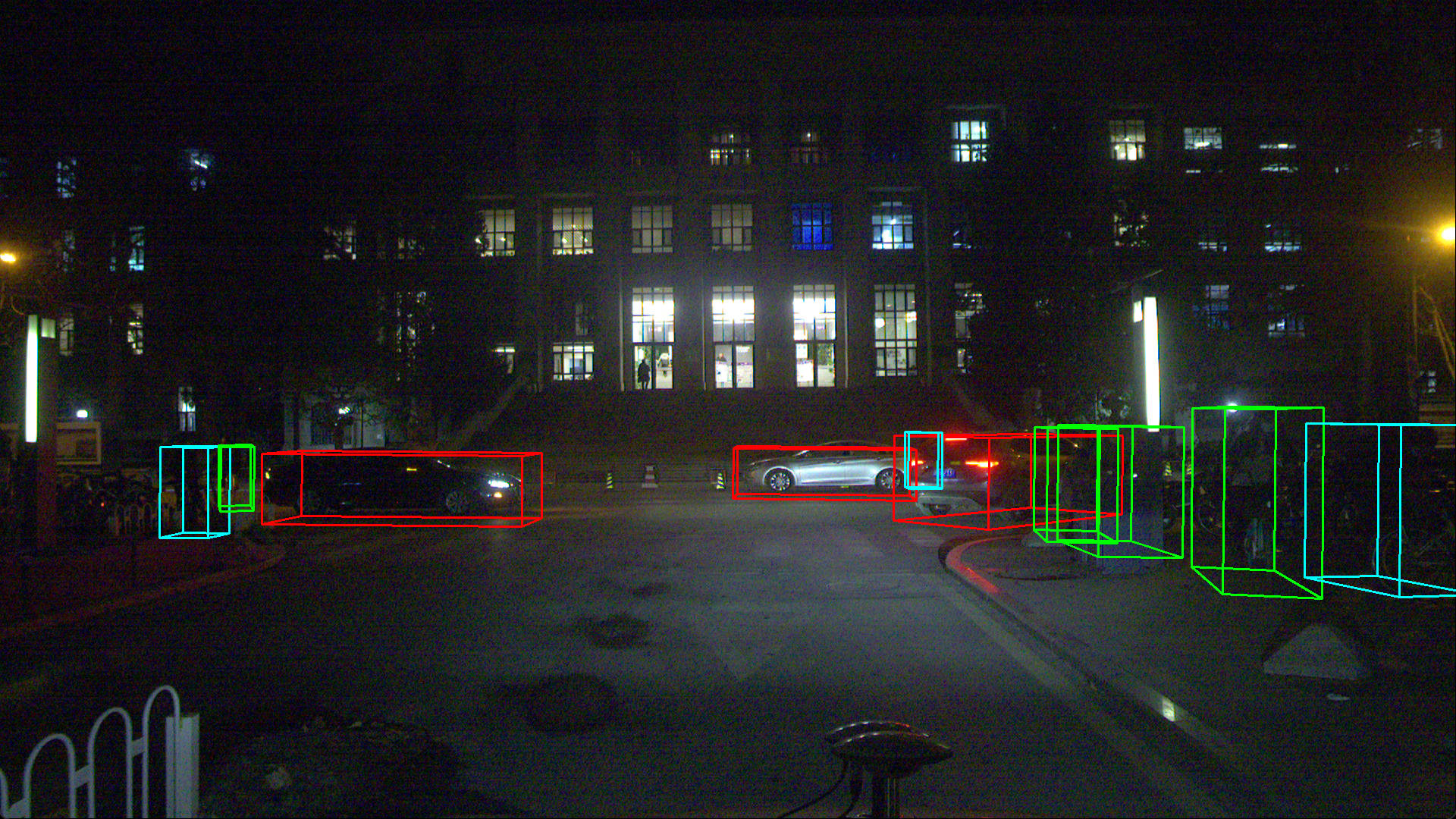}
    \end{minipage}
    \hfill
    \begin{minipage}[b]{0.3\textwidth}
        \includegraphics[width=\textwidth,height=3cm]{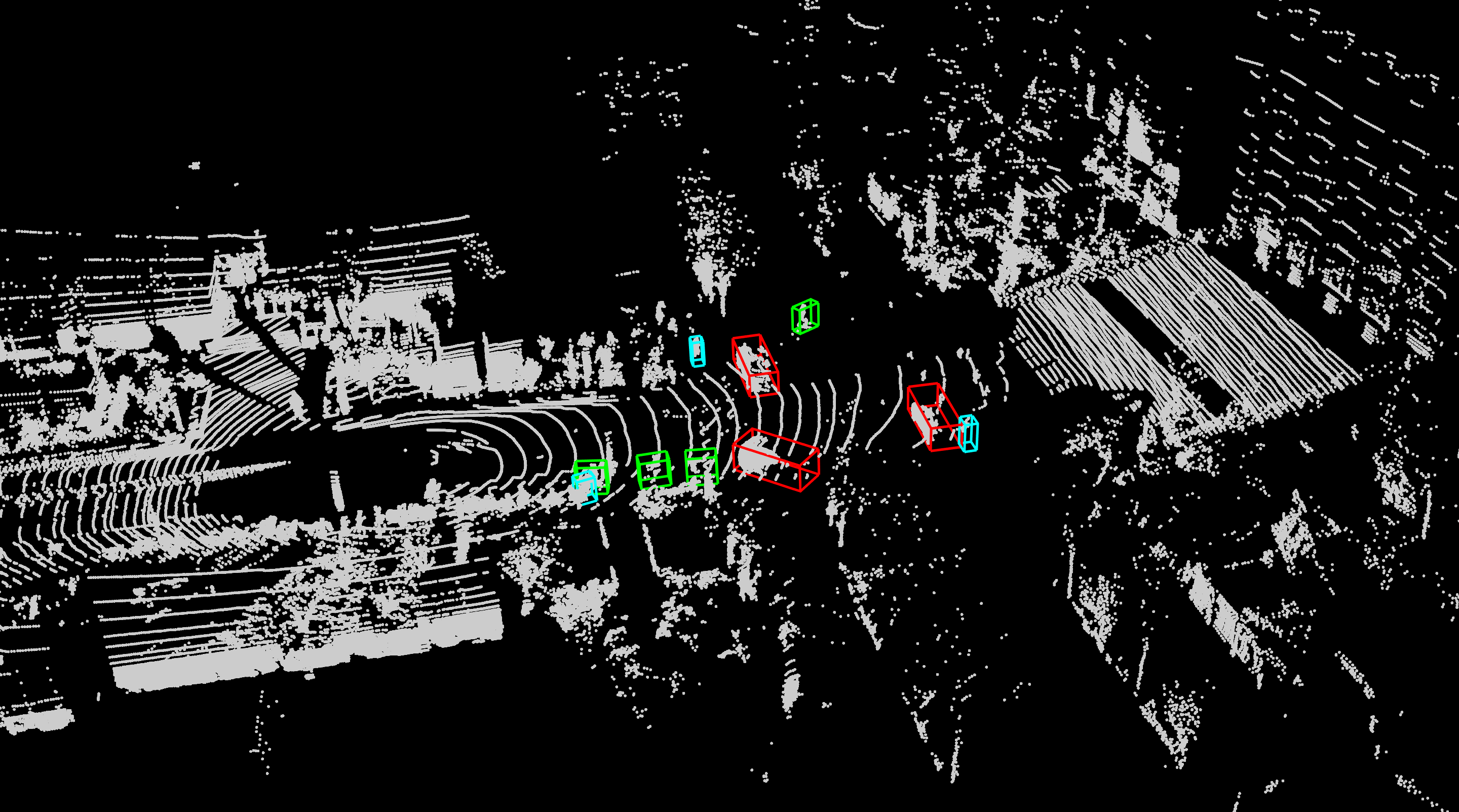}
    \end{minipage}
    \hfill
    \begin{minipage}[b]{0.3\textwidth}
        \includegraphics[width=\textwidth,height=3cm]{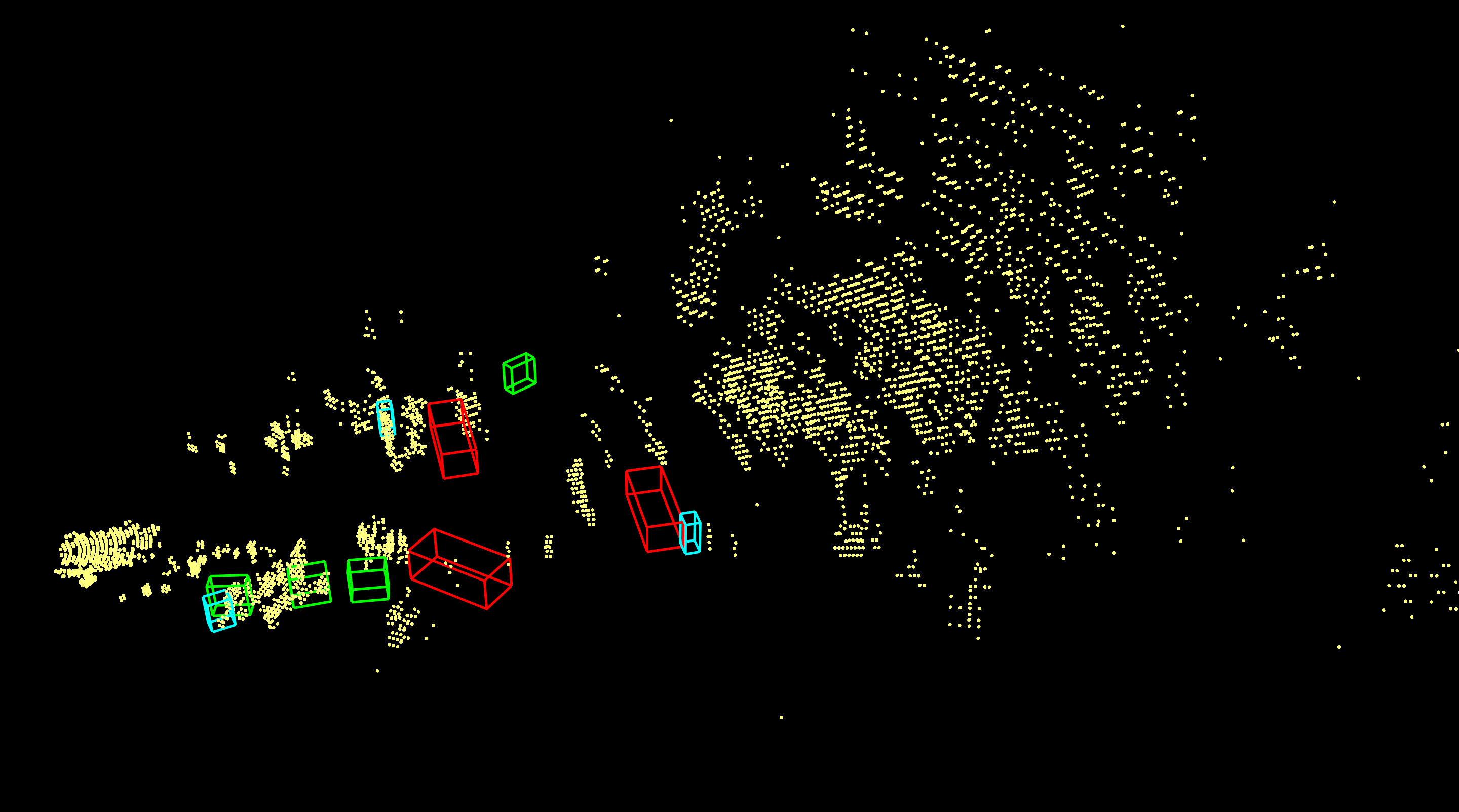}
    \end{minipage}

    \vskip\baselineskip
    \begin{minipage}[b]{0.05\textwidth}
       \footnotesize(f)
        \vspace{1.2cm}
    \end{minipage}
    \begin{minipage}[b]{0.3\textwidth}
        \includegraphics[width=\textwidth,height=3cm]{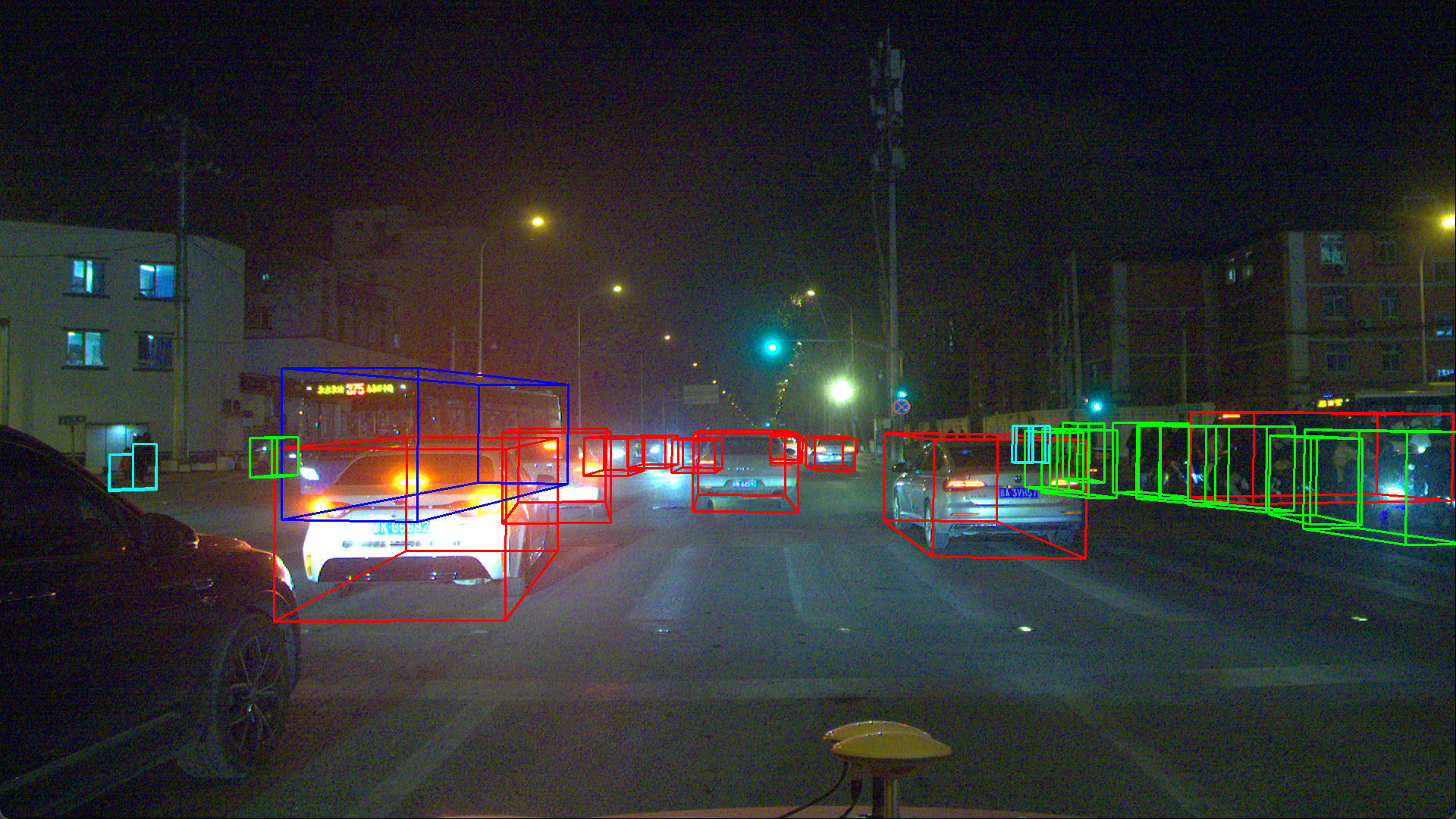}
    \end{minipage}
    \hfill
    \begin{minipage}[b]{0.3\textwidth}
        \includegraphics[width=\textwidth,height=3cm]{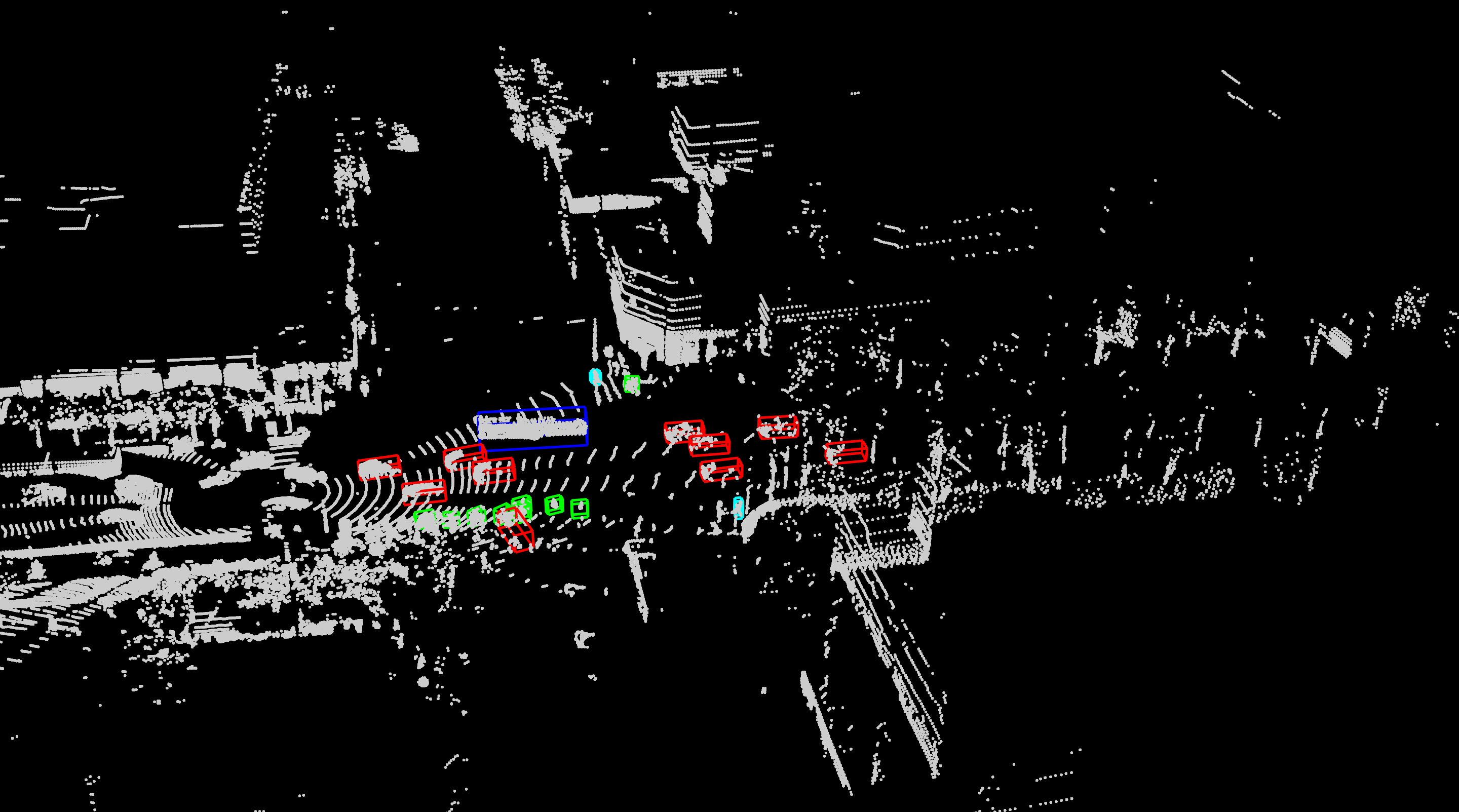}
    \end{minipage}
    \hfill
    \begin{minipage}[b]{0.3\textwidth}
        \includegraphics[width=\textwidth,height=3cm]{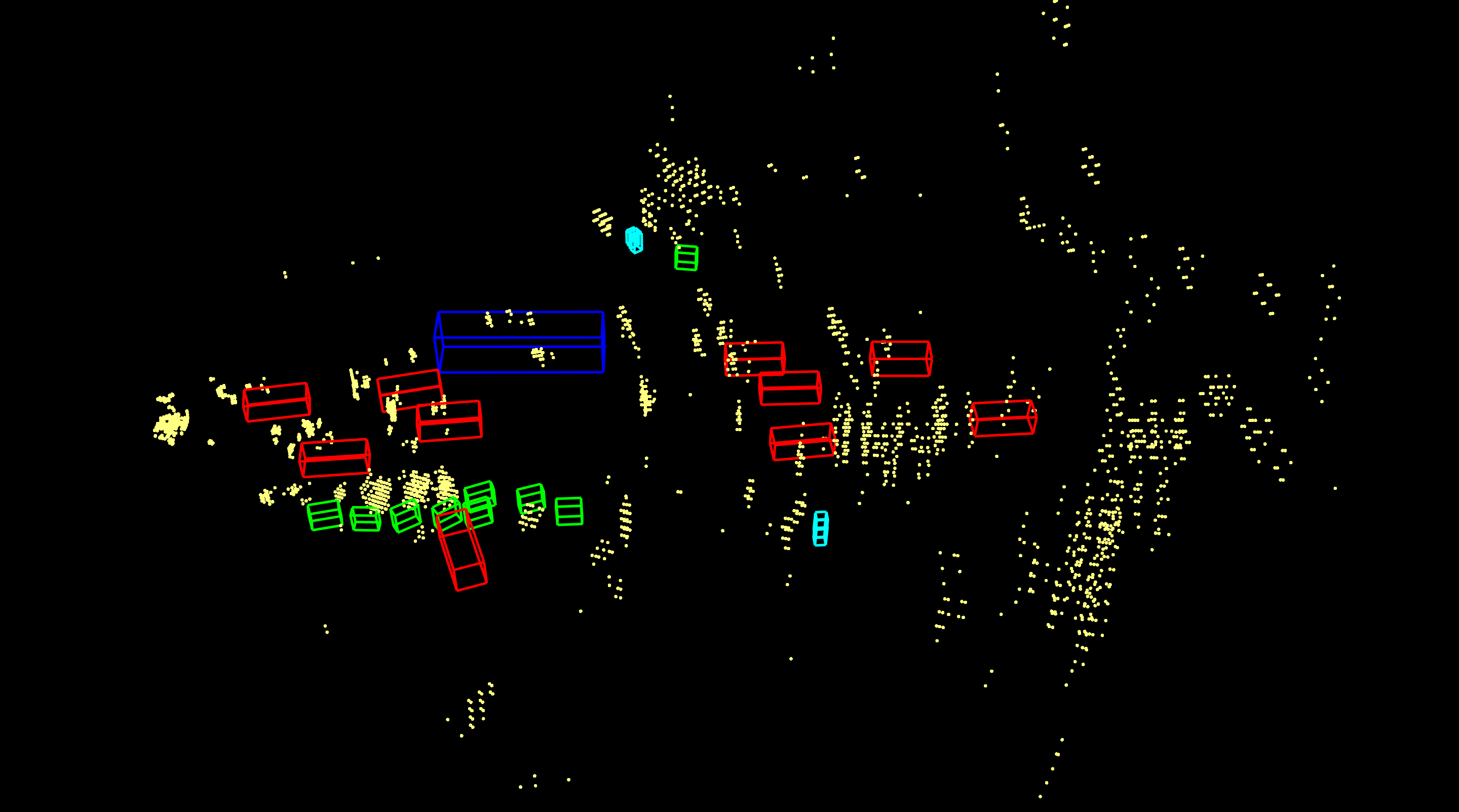}
    \end{minipage}
    \vskip\baselineskip
    \caption{Representing 3D annotations in multiple scenarios and sensor modalities. The three columns respectively display the projection of 3D annotation boxes in images, LiDAR point clouds, and LiDAR point clouds. Each row represents a scenario type. (a) Campus noon; (b) Urban noon; (c) Campus dusk; (d) Urban dusk; (e) Campus night; (f) Urban night. }
    \label{fig_vis}
    \vspace*{3\baselineskip} 
\end{figure*}

As illustrated, a visual representation of a subset of the data is presented, as shown in Fig. \ref{fig_vis}. These figures depict the results of the visualization of the labeled boxes of the multi-sensory interactive perception dataset for embodied intelligent driving proposed in this paper in two scenarios: campus and urban. Additionally, three different moments of the day are included in the urban scenario and the campus scenario: noon, afternoon, and evening. In different scenarios, the number and distribution of target objects are markedly disparate, as are the corresponding complexities of the scenarios. Additionally, the visualization results of camera, LiDAR, and millimeter-wave radar sensors are provided for each scenarioIt can be observed that though millimeter radar data is less dense than lidar data, millimeter wave radar demonstrates superior performance at long distances compared to lidar. A 3D bounding box is employed to annotate the camera image data, while a 2D bounding box is utilized to annotate the objects from the lidar and millimeter-wave radar. It is evident that the annotation boxes on the image, lidar, and millimeter-wave radar correspond well to the objects and exhibit satisfactory synchronization.

\section{EXPERIMENTS}
In this section, an experimental framework is established to empirically validate the efficacy of the dataset. M$^2$-Fusion\cite{mf} and SFD\cite{sfd} are utilized as baselines for validating the dataset, followed by a qualitative and quantitative analysis of the experimental outcomes, culminating in an evaluation of the dataset.
\subsection{Experimental Setup}
Two servers were employed for experimental validation, each running Ubuntu 20.04 and PyTorch version 1.8.1. Specifically, the SFD model underwent 40 training rounds on an Nvidia RTX3090ti graphics card with a learning rate of 0.001, while the M$^2$-Fusion model underwent 40 training rounds on an Nvidia RTX3090 graphics card with a learning rate of 0.005.
\subsection{Implementation Details}
The experimental focus lies on the "Car" target, with the dataset partitioned into training, validation, and test sets at ratios of 43.6\%, 29.7\%, and 26.7\%, respectively. This comprises 3628 training samples, 2470 evaluation samples, and 2219 test samples, totaling 8317 samples. Our datasets are available in two formats: a transformed KITTI format and a customized format dividing each continuous time series individually. Average precision (AP) serves as the performance metric, evaluated according to KITTI metrics, specifically utilizing 40-point interpolated average precision. Overlap rate thresholds are set at 50\%, and 50\% , or 70\%, and 70\% for easy, moderate, and hard difficulty targets, respectively, to assess the accuracy of detecting targets of varying difficulty with different methods.
\subsection{Quantitative Analysis}
Given the modal diversity of the dataset, 3D fusion target detection is validated using different baselines for camera, LIDAR, and millimeter-wave radar data. To establish a comprehensive understanding, the M$^2$-Fusion model serves as the baseline for fusion detection of LIDAR and millimeter-wave radar, while SFD acts as the baseline for fusion detection of camera and LIDAR. Additionally, the SFD algorithm is slightly modified for the fusion detection of camera and millimeter-wave radar, referred to as SFD-4D.

\begin{table*}[!ht]
\caption{EXPERIMENTAL RESULTS OF MULTIMODAL BASELINES FOR DIFFERENT DIFFICULTY LEVELS\label{tab:baselines}}
\centering
\setlength{\tabcolsep}{17pt}
\renewcommand{\arraystretch}{1.5}
\begin{tabular}{c c c c c c c c}
\toprule[2px]

\textbf{Baselines} & \textbf{Data} & \textbf{Mode} & \textbf{Easy} & \textbf{Mod} & \textbf{Hard} & \textbf{Recall} & \textbf{Eval Time/ms} \\
\hline
\hline
\multicolumn{8}{c}{\textbf{DECECTION RESULT AT AOI OF 0.7, and 0.7}}\\
\hline
\hline
 \multirow{2}{*}{M$^2$-Fusion\cite{mf}} & \multirow{2}{*}{Lidar + Radar} & 3D &68.50 &44.07&37.74 & \multirow{2}{*}{33.34}  & \multirow{2}{*}{158.70} \\

                              &                                 & BEV &83.60 &55.36&48.31& \multirow{2}{*}{}  & \multirow{2}{*}{}  \\

\multirow{2}{*}{SFD\cite{sfd}}          & \multirow{2}{*}{Camera + Lidar} & 3D &67.85 &48.88 &43.05 & \multirow{2}{*}{45.59}  & \multirow{2}{*}{35.10} \\

                              &                                 & BEV &68.62 &48.55 &43.58& \multirow{2}{*}{}  & \multirow{2}{*}{} \\

\multirow{2}{*}{SFD-4D}       & \multirow{2}{*}{Camera + Radar} & 3D &33.71 &19.57 &14.88 &\multirow{2}{*}{16.30}  & \multirow{2}{*}{41.10} \\

                              &                                 & BEV &52.41 &31.71 &24.38 &\multirow{2}{*}{}  & \multirow{2}{*}{} \\
\hline
\hline
\multicolumn{8}{c}{\textbf{DECECTION RESULT AT AOI OF 0.5, and 0.5}}\\
\hline
\hline
\multirow{2}{*}{M$^2$-Fusion\cite{mf}} & \multirow{2}{*}{Lidar + Radar} & 3D &88.14 &59.36&53.87 & \multirow{2}{*}{33.34}  & \multirow{2}{*}{158.70} \\

                              &                                 & BEV &88.79 &61.36&54.43& \multirow{2}{*}{}  & \multirow{2}{*}{}  \\

\multirow{2}{*}{SFD\cite{sfd}}          & \multirow{2}{*}{Camera + Lidar} & 3D &75.53 &49.48 &44.74 & \multirow{2}{*}{45.59}  & \multirow{2}{*}{35.10} \\

                              &                                 & BEV &75.76 &49.54 &46.39& \multirow{2}{*}{}  & \multirow{2}{*}{} \\

\multirow{2}{*}{SFD-4D}       & \multirow{2}{*}{Camera + Radar} & 3D &63.84 &37.28 &26.98 &\multirow{2}{*}{16.30}  & \multirow{2}{*}{41.10} \\

                              &                                 & BEV &66.29 &41.25 &30.71 &\multirow{2}{*}{}  & \multirow{2}{*}{} \\

\bottomrule[2px]
\end{tabular}

\end{table*}

At an AOI threshold of 0.7, as can be seen from the Table \ref{tab:baselines}, due to the fact that our dataset is set up with many challenging scenarios for different lighting conditions, the confidence of the camera data is greatly affected, which results in the detection accuracy of the camera fused with LIDAR and millimeter-wave radar for a simple target in BEV view of only 68.62\% and 52.41\%, respectively. In contrast, since LIDAR and LiDAR are not affected by lighting conditions, the detection accuracy reaches 83.60\%, which is 14.98\% and 31.19\% higher than that of the former two, respectively. It also proves that the LiDAR and millimeter wave radar data we collected can play a good complementary role when in the night or in scenes such as drastic changes in lighting.

Analyzing further, we can find that in recent years, 3D target detection algorithms have gradually adopted a new fusion strategy, i.e., the camera data is transformed into a pseudo-point cloud by monocular depth estimation and projection transformation, and then fused with the point cloud collected by lidar or radar. Although this algorithm can ensure a faster detection rate (the detection rate can reach more than 20Hz, while the detection rate of M$^2$-Fusion is around 7Hz) and maintain a higher detection accuracy in most of the scenes, the pseudo-point cloud transformed by the camera can be fused into a pseudo-point cloud by monocular depth estimation and projection transformation, and then fused with the point cloud collected by lidar or radar. However, due to the huge amount of information in the pseudo-point cloud data transformed by the camera, the detection result obtained by the algorithm depends largely on the quality of the pseudo-point cloud without targeted processing. Therefore the detection results will drop dramatically as the quality of the camera data decreases.

In addition, from the table, we can find that when the AOI settings are 0.5, the detection results of camera-lidar fusion are 9.47\% higher than those of camera-radar fusion for the simple difficulty "car" target in BEV view, and 8.29\% and 15.68\% higher for the medium and difficult difficulties, respectively. In contrast, when the AOI settings were elevated to 0.7, the detection results of SFD-4D decreased dramatically by 13.88\%, 9.54\% ,and 6.33\%, while those of SFD decreased only by 7.14\%, 0.99\% and 2.81\%. This is due to the fact that the 80-line LiDAR has a higher point density and collects much more information than the LiDAR, resulting in higher detection accuracy and more accurate prediction frames.

Based on these findings, we modified the SFD model slightly to introduce light and vibration data in order to further explore the effects of environmental conditions on detection performance.  We selected data from the afternoon time period for our experiments (2,300 frames), when the effects of dynamic changes in light and vibration conditions were strongest, as shown in Table \ref{tab:light_vibration}. The results show that the detection model after fusing the light and vibration data improves the detection accuracy in the BEV view by 4.09\%, 3.01\%, and 3.79\%, at an AOI of 0.5. Similar improvements were observed for all difficulty levels at an AOI of 0.7. Moreover, the introduction of light and vibration data barely affected the detection efficiency of the model. These consistent improvements coupled with stable execution times highlight the importance of light and vibration data for perceptual models.

\begin{table*}[!ht]
\caption{EXPERIMENTAL RESULTS WITH AND WITHOUT LIGHT/VIBRATION EFFECTS\label{tab:light_vibration}}
\centering
\setlength{\tabcolsep}{15pt}
\renewcommand{\arraystretch}{1.5}
\begin{tabular}{c c c c c c c}
\toprule[2px]

\textbf{Condition} & \textbf{Mode} & \textbf{Easy} & \textbf{Mod} & \textbf{Hard} & \textbf{Recall} & \textbf{Eval Time/ms}\\
\hline
\hline
\multicolumn{7}{c}{\textbf{DETECTION RESULT AT AOI OF 0.7, and 0.7}} \\
\hline
\hline
\multirow{2}{*}{With Light/Vibration} & 3D & 55.60 & 41.14 & 41.26 & \multirow{2}{*}{46.04} & \multirow{2}{*}{57.4} \\
                                      & BEV & 55.88 & 41.41 & 41.48 &  &  \\
\multirow{2}{*}{Without Light/Vibration} & 3D & 51.51 & 38.13 & 37.47 & \multirow{2}{*}{45.08} & \multirow{2}{*}{59.3} \\
                                         & BEV & 51.85 & 38.42 & 37.89 &  &  \\
\hline
\hline
\multicolumn{7}{c}{\textbf{DETECTION RESULT AT AOI OF 0.5, and 0.5}} \\
\hline
\hline
\multirow{2}{*}{With Light/Vibration} & 3D & 56.29 & 41.76 & 41.78 & \multirow{2}{*}{46.04} & \multirow{2}{*}{57.4} \\
                                      & BEV & 57.62 & 41.94 & 41.98 &  &  \\
\multirow{2}{*}{Without Light/Vibration} & 3D & 51.06 & 39.11 & 38.53 & \multirow{2}{*}{45.08} & \multirow{2}{*}{59.3} \\
                                         & BEV & 53.23 & 39.46 & 39.00 &  &  \\
\bottomrule[2px]
\end{tabular}
\end{table*}

\section{DISCUSSION AND FUTURE WORK}
The experimental results demonstrate that existing fusion detection algorithms require further improvement in terms of accuracy when confronted with challenging scenarios, such as poor light or uneven road surfaces. This issue also presents a significant challenge for current automatic driving perception algorithms. The light and vibration data provided in our dataset offer a potential avenue for subsequent algorithm research. By evaluating the environment through light, vibration, sound, and other environmental state data, the fusion detection algorithm can be dynamically adjusted according to different conditions. This allows us to better utilize the modal data and improve the algorithm's generalizability.

This dataset will also provide strong data support for embodied intelligent driving and open a new path for the development of automatic driving. The environmental state information, such as light and vibration, collected by embodied intelligent vehicles can be used to obtain more accurate and effective embodied semantic information and to perceive the surrounding environment more comprehensively. Additionally, our data acquisition platform has been expanded based on the Dual Radar dataset collection platform. Consequently, the camera, lidar, and Arbe Radar data that have been collected can supplement the Dual Radar dataset, thereby enhancing algorithmic performance.

In the future, an expanded dataset will be constructed based on the current one, incorporating additional modalities such as vehicle speed and trajectories. This enhanced dataset will include more complex scenarios to support end-to-end tasks from perception to path planning and decision-making, thereby advancing research in embodied intelligent driving systems. Furthermore, new algorithms will be developed to enhance the utilization of environmental state data for embodied intelligent driving.

\section{CONCLUSION}
A multi-sensory interactive perception dataset for supporting research in embodied intelligent driving is presented, which contains light, vibration, sound, and IMU data in addition to camera, LiDAR, and millimeter-wave radar data. This dataset can be used for 3D object detection, target tracking, and future embodied sensing tasks in autonomous driving. To ensure the richness of the light, vibration, and sound data, a series of challenging scenarios were designed. These scenarios also serve to evaluate the performance of the algorithms under different conditions. Two baselines were employed to test the dataset, and the results demonstrated that it is capable of meeting the current autopilot perception task requirements. Furthermore, the scenario settings of the dataset present novel challenges for the autopilot fusion perception algorithm.

\section*{Acknowledgments}
We thank Datatang (www.datatang.ai) company for providing dataset annotation services.

\newpage

\vspace{11pt}

\vspace{11pt}

\vfill

\end{document}